\documentclass[10pt,journal]{IEEEtran}

\usepackage{color}
\usepackage{tabularx}
\usepackage{bm}
\usepackage{multirow}
\usepackage{amsmath}
\usepackage{amsfonts}
\usepackage{siunitx}
\usepackage{graphicx}
\usepackage{subfigure}
\usepackage{threeparttable}
\usepackage{enumerate}
\usepackage{subfigure}
\usepackage{enumerate}
\usepackage{subfigure}
\usepackage{algorithm}
\usepackage{algorithmicx}
\usepackage{algpseudocode}
\usepackage{tablefootnote}
\usepackage{pifont}
\newtheorem{remark}{Remark}

\ifCLASSOPTIONcompsoc
  \usepackage[nocompress]{cite}
\else
  \usepackage{cite}
\fi

\ifCLASSINFOpdf
\else
\fi

\begin{document}
%
\title{\huge{Adversarial RAW: Image-Scaling Attack Against Imaging Pipeline}}
\author{Junjian~Li,~Honglong~Chen,~{\em Senior~Member,~IEEE}}

\markboth{}{}

\IEEEtitleabstractindextext{%
\begin{abstract}
  Deep learning technologies have become the backbone for the development of computer vision. With further explorations, deep neural networks have been found vulnerable to well-designed adversarial attacks. Most of the vision devices are equipped with image signal processing (ISP) pipeline to implement RAW-to-RGB transformations and embedded into data preprocessing module for efficient image processing. Actually, ISP pipeline can introduce adversarial behaviors to post-capture images while data preprocessing may destroy attack patterns. However, none of the existing adversarial attacks takes into account the impacts of both ISP pipeline and data preprocessing. In this paper, we develop an image-scaling attack targeting on ISP pipeline, where the crafted adversarial RAW can be transformed into attack image that presents entirely different appearance once being scaled to a specific-size image. We first consider the gradient-available ISP pipeline, i.e., the gradient information can be directly used in the generation process of adversarial RAW to launch the attack. To make the adversarial attack more applicable, we further consider the gradient-unavailable ISP pipeline, in which a proxy model that well learns the RAW-to-RGB transformations is proposed as the gradient oracles. Extensive experiments show that the proposed adversarial attacks can craft adversarial RAW data against the target ISP pipelines with high attack rates.
\end{abstract}

}

\maketitle
\renewcommand{\thefootnote}{\fnsymbol{footnote}}
\setcounter{footnote}{0}


\IEEEdisplaynontitleabstractindextext

%
\IEEEpeerreviewmaketitle

\section{Introduction}\label{sec:introduction}
Deep learning has obtained or even exceeded human-level performance in many vital fields, such as computer vision~\cite{image1,image2}, natural language processing~\cite{NLP1,NLP2}, medical diagnosis~\cite{med1,med2} and so on. The wide demand in the computer vision domain is one of the driving forces for the sustainable development of deep learning. Deep neural networks (DNNs), as the cornerstone technology in deep learning, are gradually becoming the preference for complicated tasks and widely applied in diverse vision applications.

A mass of high-risk vision applications make the decisions based on the outputs of DNNs, but the reliability of model outputs can be critical. However, existing results have shown that DNNs are vulnerable to adversarial attacks, which can skew model predictions or expose data privacy. Adversarial attacks undoubtedly create the potential threats for the applications of DNNs in security-sensitive fields. Typical examples of the adversarial attacks include adversarial examples~\cite{ori_adv}, backdoor attacks~\cite{backdoor} and inferring properties attacks~\cite{infer1}. Although researchers have done a great deal of work for these attacks, the attacks on the vulnerabilities of DNNs are still being investigated.

\begin{figure}[htbp]
	\centering
	\includegraphics[height=4.03cm,width=8.5cm]{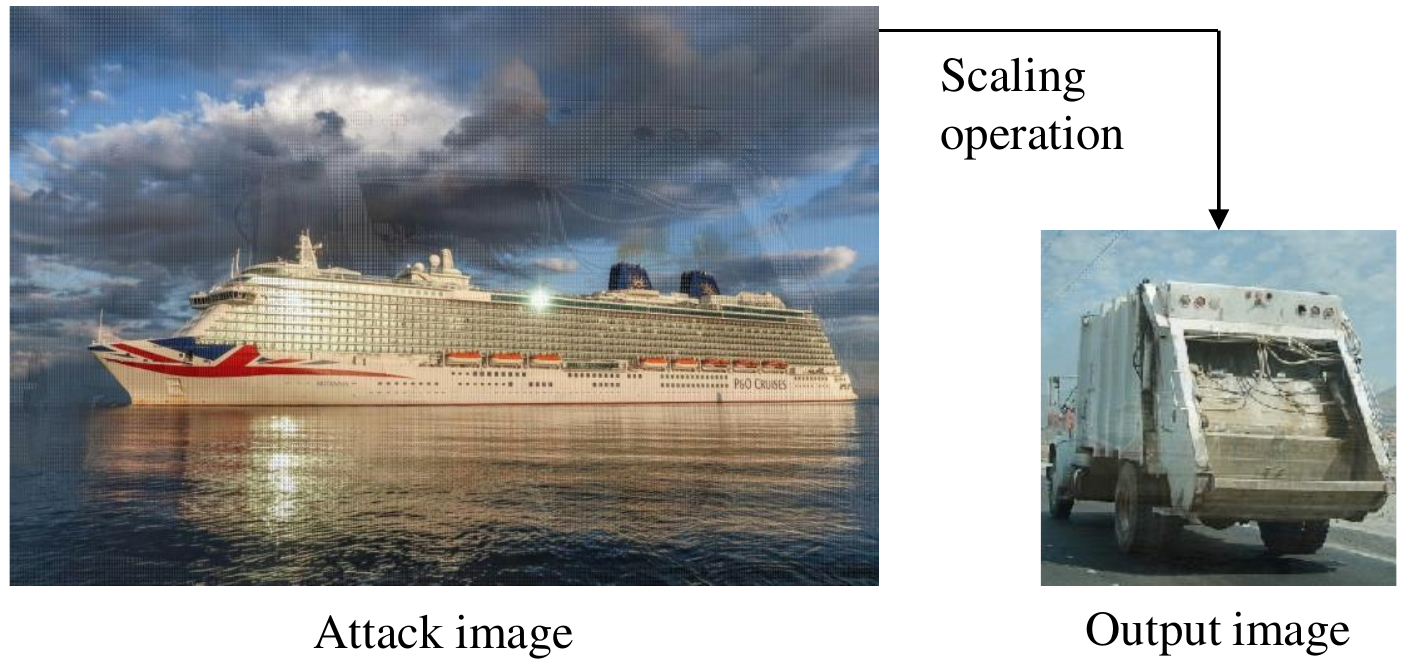}
	\caption{An example of the image-scaling attack, where a ferry is scaled to a truck.}
	\label{attack_example}
\end{figure}

The fix-size input layers of DNNs are commonly defined. Since most provided images from diverse sources vary in sizes, DNNs based vision models are configured with a preprocessing program, called image scaling, to reconstruct images of fit sizes while preserving their initial `semantics'. However, the weakness of the preprocessing procedure is always ignored. Recently, Xiao~\textit{et al.}~\cite{xiao2019} explored the vulnerability of the data preprocessing and proposed an attack against image scaling operation, called image-scaling attack. Through introducing well-designed perturbations, image-scaling attack can craft adversarial images presenting entirely different appearances once scaled to specific sizes. Fig.~\ref{attack_example} presents an example of the image-scaling attack. As with existing adversarial attacks, image-scaling attack can also be used as the data poisoning means to disturb the training and inference procedures of models. Differently, image-scaling attack is independent of certain models or features and coexists with the preprocessing operation, which should be well taken into consideration in the deep learning pipeline.

\begin{figure}[h]
	\centering
	\includegraphics[height=2.89cm,width=8.3cm]{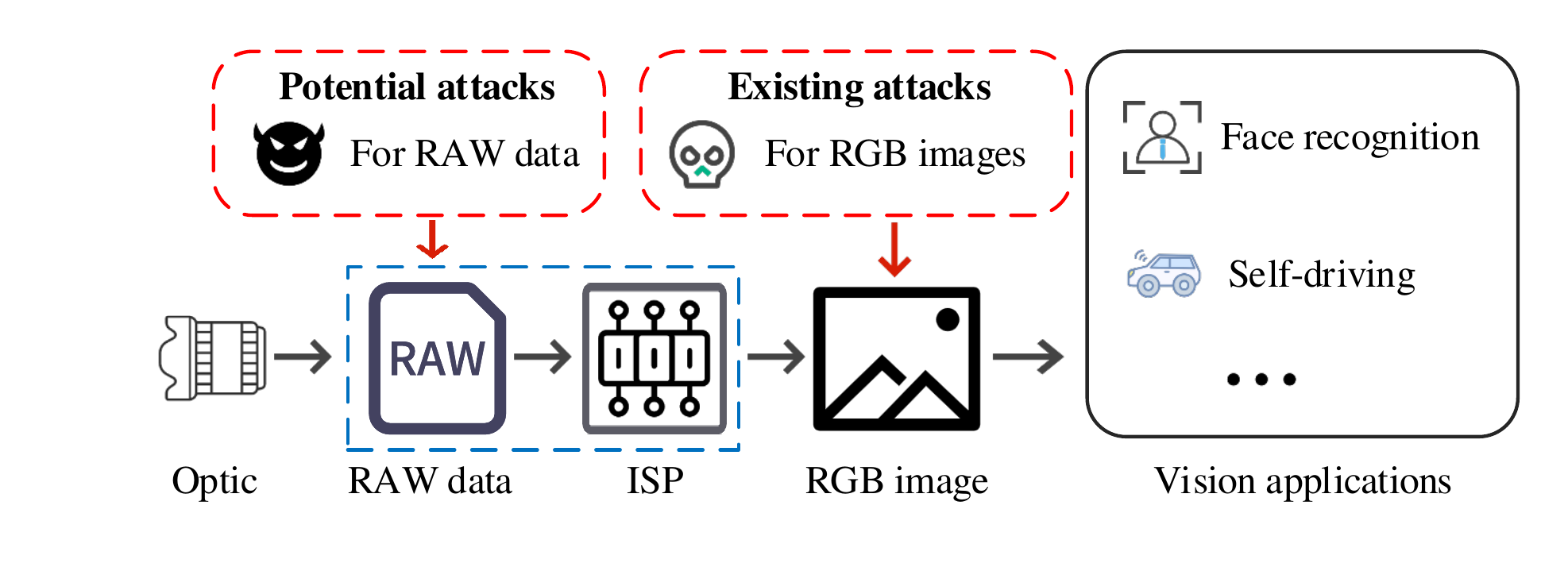}
	\caption{Existing adversarial attacks against vision applications. Most adversarial attacks aim to manipulate the post-capture images while neglecting potential adversarial patterns introduced by ISP pipeline.}
	\label{attack_status}
\end{figure}

The aforementioned adversarial attacks aim at tampering with the post-capture images, i.e., the RGB images, but they all ignore an essential intermediate procedure: \textit{image signal processing (ISP) pipeline.} Existing vision devices, such as smartphones and self-driving automobiles, are configured with ISP modules to transform RAW signals captured by imaging sensors into RGB images through implementing intricate operations. Actually, the quality of RGB images relies on RAW data. Malicious behaviors on RAW data may be involved into RGB images through ISP procedure. Most of the adversarial attacks and defense mechanisms pay more attention on the post-capture images, as shown in Fig.~\ref{attack_status}. However, the unprocessed RAW data is not suitable for human eyes, making the introduced perturbations on RAW data more inconspicuous. More importantly, the conventional view is that ISP is secure and attacks are only for RGB images, which can cause malicious operations on RAW to be ignored. Thus, it motivates us to explore the adversarial attacks on the ISP procedure in this paper.

Although the prior work~\cite{adv_ISP} designed the adversarial attacks against ISP procedure, it assumed that deep models directly consume the obtained adversarial images without considering the preprocessing process. The preprocessing operation, specifically image scaling, can destroy adversarial patterns of attack images~\cite{guo2017countering}. Notably, most of the vision devices in reality are equipped with ISP pipeline to implement RAW-to-RGB transformations and embedded into data preprocessing module for efficient image processing. Nevertheless, existing adversarial attacks ignore the impacts of either the ISP pipeline or the scaling operation. A lack of comprehensive consideration can lead to the omissions of weaknesses in vision application models. In this paper, different from the general adversarial attacks, we focus on perturbing RAW data before ISP procedures to generate attack images that can be against the scaling operation. Specifically, the crafted RAW data can be transformed by ISP pipeline into the attack images that present completely different appearances once being scaled. We first consider the gradient-available ISP, in which the gradient information can be directly used in generating the adversarial RAW. To make the adversarial attack more applicable, we further consider the gradient-unavailable ISP, and the basic idea behind is to approximate the target ISP pipeline via deep learning as the proxy model, which can well learn the RAW-to-RGB transformations. Then the image-scaling attack against the target ISP pipeline can be launched by utilizing the approximated gradient information from the proxy model.

The key contributions of this paper are summarized as:
\begin{itemize}
    \item We develop an image-scaling attack targeting on ISP pipeline, which considers the impacts of both the ISP pipeline and data preprocessing in vision applications.
	\item We first propose the gradient-available ISP attack, in which the gradient information of the ISP pipeline is directly used in generating the adversarial RAW.
	\item We further propose the gradient-unavailable ISP attack, in which a proxy model is designed as the gradient oracles by well learning the RAW-to-RGB transformations.
	\item We implement extensive experiments to validate the effectiveness of the proposed attacks, which can craft adversarial RAW data against the ISP pipelines with high attack rates.
\end{itemize}

The rest of this paper is organized as follows. Section~\ref{sec:related_work} reviews the related work. The preliminaries of ISP pipeline and image-scaling attack are presented in Section~\ref{sec:preliminaries}. Section~\ref{sec:attack} illustrates the designed attack mechanism. Section~\ref{sec:experiment} gives the experimental results and analysis. The defense methods are discussed in Section~\ref{sec:defense}. Finally, Section~\ref{sec:conclusion} presents the conclusions and future works.

\section{Related Work}\label{sec:related_work}
In this section, we briefly review some topics related to our work.
 \subsection{Image Signal Processing}
 ISP pipeline reconstructs RGB images suitable for human eyes from RAW sensor signals. There are many researches targeting on the convention ISP pipeline. Heide~\textit{et al.}~\cite{flexisp} incorporated with the traditional image processing procedures to improve the quality of reconstructed images. In~\cite{burst}, a computational imaging pipeline for smartphones was proposed to reduce noise and intensify dynamic ranges. Xu~\textit{et al.}~\cite{color_trans} proposed a spatially-variant color transformation means to implement more efficient color corrections in ISP pipeline. However, for the conventional ISP pipeline, each processing module is usually exploited and rectified independently without considering the influences of the successors, which may result in deviation accumulations in the processing procedure. Moreover, the conventional ISP pipeline cannot cope with the imaging demands along with the universal smartphone photography due to the simple algorithms embedded into each module. Thus, some learning-based ISP methods are proposed. Jiang~\textit{et al.}~\cite{learning_isp} utilized clustered RAW patches based on simple features and learned a mapping from RAW data to RGB images. In~\cite{learning_dark}, a multiscale CNN model was trained for low-light image processing, which can transform the noisy RAW data to clean RGB images. Schwartz~\textit{et al.}~\cite{deepisp} presented DeepISP to be a proxy as the the camera image signal processing pipeline, which can implement low-level tasks, such as demosaicing and denoising, as well as higher-level tasks, such as color correction and image adjustment. Note that the mentioned above learning-based approaches are task-specific and not general for various complex scenes. Liang~\textit{et al.}~\cite{cameranet} proposed CameraNet model for multiple tasks inside an ISP pipeline. CameraNet consists of two relatively uncorrelated optimization modules of an ISP pipeline: restoration and enhancement. Optimization of ISP pipeline is an ongoing topic.
 \subsection{Typical Adversarial Attacks}
 A mass of studies explored adversarial attacks against deep learning models in computer vision. Szegedy~\textit{et al.}~\cite{ori_adv} firstly crafted images with subtle perturbations to deceive deep learning models, called adversarial examples. Adversarial examples can result in the incorrect predictions of models. Henceforth, the results of adversarial examples have continued to emerge. In~\cite{fgsm}, Goodfellow~\textit{et al.} argued that adversarial examples result from high-dimensional linearity and proposed an attack method called Fast Gradient Signed Method. Common attack means, such as DeepFool~\cite{deepfool} and C$\&$W~\cite{cw} and so on, are considered to be white-box settings, i.e., an adversary is fully aware of the properties of the target models. Moreover, there are some generating methods targeting on black-box scenarios, such as MI-FGSM~\cite{mifgsm}, Curls$\&$Whey~\cite{curls} and one pixel attack~\cite{onepix}.

 Differently, Liu~\textit{et al.}~\cite{backdoor} proposed a trojaning attack on deep neuron networks, called backdoor attack. It utilizes the trojan triggers to inject malicious behaviors into models, and those behaviors are only aroused by inputs with the trojan triggers. The proposed attack mechanism is called the patch-based method, which focuses on trojaning with non-semantic patches. Instead, Lin~\textit{et al.}~\cite{comp_backdoor} introduced composite backdoor attack to uses composition of existing benign features or objects as the trigger, which is more stealthy and can elude backdoor scanners. Interestingly, different from the attack pattern of modifying training data, a blind backdoor attack based on compromising the loss-value computation in the model-training code was investigated in~\cite{blind_backdoor}.

 Moreover, researchers also have investigated the leakage of model properties. In~\cite{infer1}, the membership inference (MI) was proposed, which aims to infer whether a specific data belongs to the training set of the target model. Hui~\textit{et al.}~\cite{blind_mi} utilized differential comparison approach to probe the target model and extract membership semantics. Departing from attacks on classification models, results of membership inference targeting on generative models~\cite{mi_gan} or federated learning~\cite{mi_fl} were also proposed.

 Due to the complexity and inexplicability of deep learning models, more potential vulnerabilities still need to be explored.

\section{Preliminaries}\label{sec:preliminaries}
This section gives the basic knowledge of ISP pipeline and image-scaling attack, respectively.
 \subsection{ISP Pipeline}
 The digital camera is embedded with multiple modules, which constitute the ISP pipeline, to transform RAW sensor data into RGB images. Commonly, each process in ISP pipeline requires perpetual experimental tuning for specific scenarios or cameras. Typical ISP pipeline generally involves the following procedures:
 \begin{enumerate}[(1)]
    \item \textbf{Optics:} Firstly, the common optic systems focus the light fields on a set of photodiodes, which can obtain RAW digital values through an analog-to-digital conversion circuit.
    \item \textbf{White balance $\&$ gain:} After the black level bias removes and defect pixel corrections, the values of RAW data are color-corrected and gain-adopted on the basis of illuminant colors~\cite{white_balance}.
    \item \textbf{Demosaicking:} RGB values are extracted from RAW pixels, for example by applying the interpolation method~\cite{demosaicking}.
    \item \textbf{Denoising:} The weak sensor noises in the obtained RGB values are eliminated using some filtering methods, such as the non-local patch mapping and the edge-preserving filtering.
    \item \textbf{Color $\&$ Tone Correction:} Some further corrections are used to improve the image quality, such as the global manipulations (the gamma curve) or local manipulations (sharpening).
    \item \textbf{Colorspace Conversion $\&$ Compression:} Pixel values are transformed into the specific colorspace before compression, storage, or further processing.
 \end{enumerate}

 Note that, most adversarial attacks aim at manipulating the RGB images without considering the ISP pipeline or assume that the ISP pipeline is secure~\cite{ori_adv,deepfool,backdoor,comp_backdoor,physical_attack}. However, most of the vision devices are configured with ISP modules to implement RAW-to-RGB transformations. Unfortunately, ISP pipeline can also introduce malicious behaviors to the RGB images. Thus, the adversarial attacks on the ISP pipeline should be explored.
\subsection{Image-Scaling Attack}
\begin{figure}[htbp!]
	\centering
	\includegraphics[height=3.0cm,width=5.8cm]{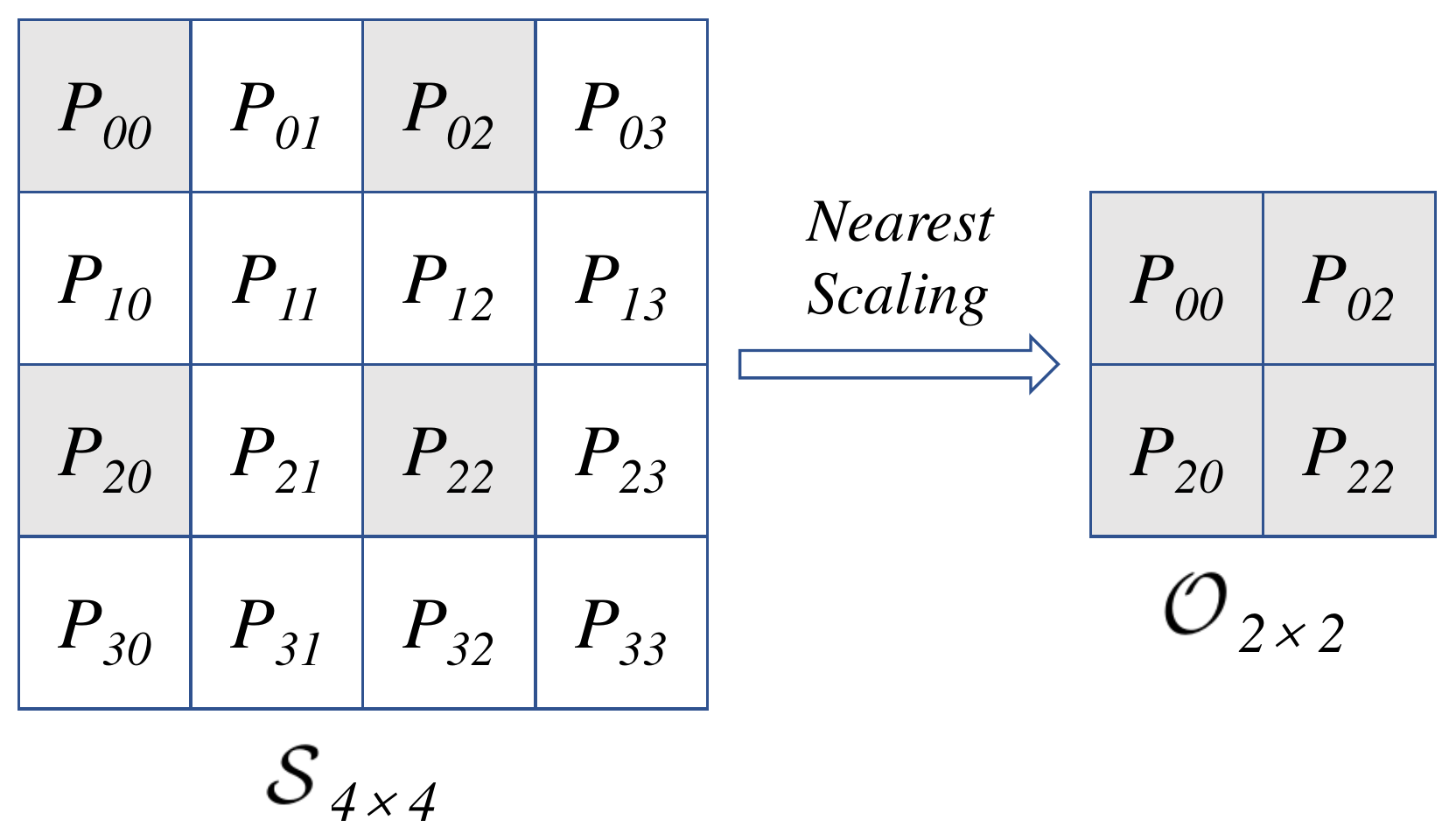}
	\caption{An example of image scaling using Nearest algorithm, where most pixels (white blocks) are discarded while others (shaded blocks) are preserved to form the contents after scaling.}
	\label{scaling_example}
\end{figure}
The image scaling operation is an essential procedure to adjust images of different sizes in a typical deep learning pipeline. Vision applications based on deep learning models often require the fixed size inputs while the obtained images vary in dimensions. Thus, to make images match the sizes expected by models, scaling operation becomes the necessary procedure before image processing. Image scaling can be considered as a `signal resampling' procedure, which only adjusts the size of image while maintaining its initial `semantics'~\cite{scaling_operation}. The most commonly used scaling algorithms are Nearest, Bilinear and Bicubic. One example of the scaling process is illustrated in Fig.~\ref{scaling_example}, i.e., some pixels (shaded blocks) are remained (or extracted to implement computational process) and others (white blocks) are discarded when scaling.

\begin{figure}[htbp!]
	\centering
	\includegraphics[height=3.47cm,width=8.2cm]{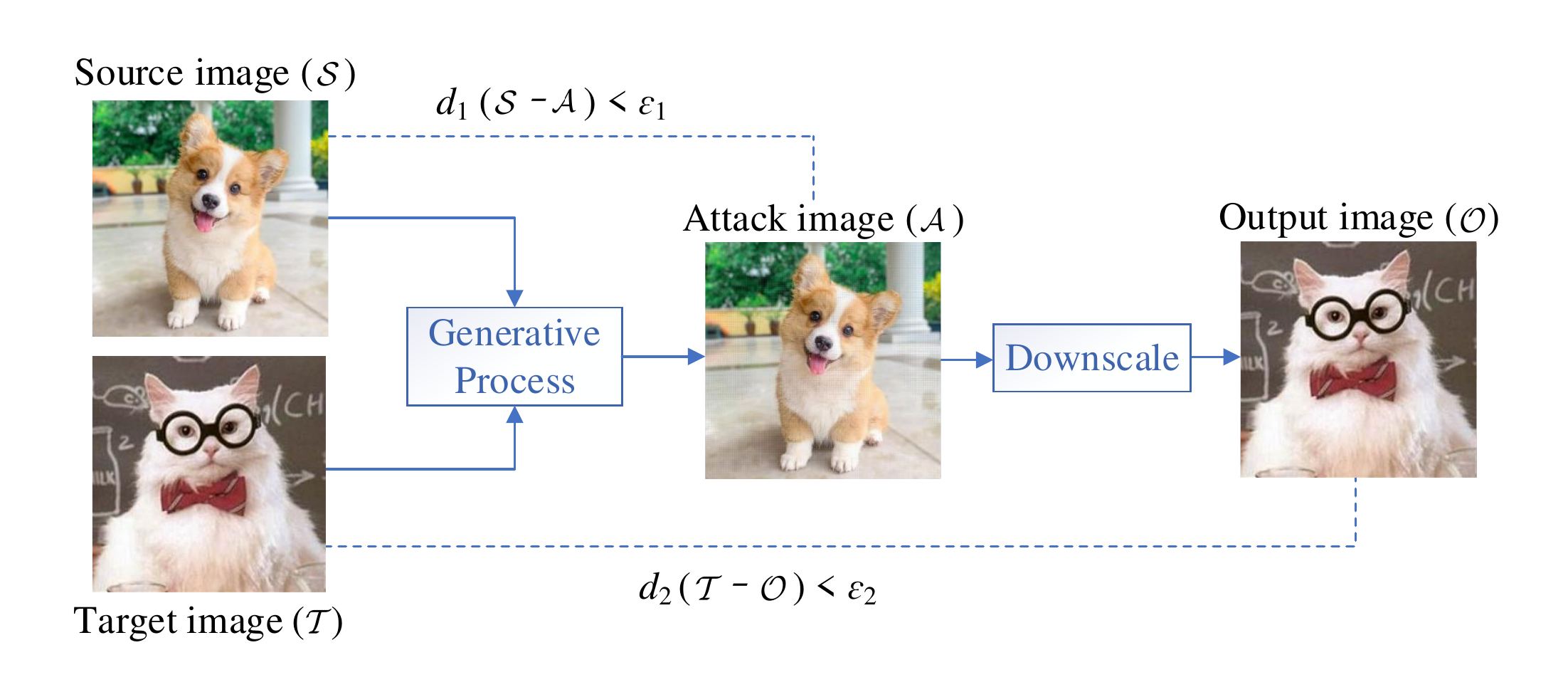}
	\caption{The process of crafting an attack image. $\varepsilon_{1}$ and $\varepsilon_{2}$ are two small thresholds. The dimension of the output (target) image is smaller than that of the attack (source) image, we present them as the same dimension for the sake of composition.}
	\label{scaling_attack}
\end{figure}

Recently, Xiao~\textit{et al.}~\cite{xiao2019} proposed the image-scaling attack against the scaling operation, which aims to craft an attack image that becomes the completely different one after scaling. Actually, considering the example in Fig.~\ref{scaling_example}, an adversary can tamper with the final remained or extracted pixels and render others unchanged to launch an image-scaling attack. Ultimately, the modified pixels form the target contents after scaling. Specifically, a source image $\mathcal{S}$ can be modified to an attack image $\mathcal{A}$ that resembles a target image $\mathcal{T}$ after downscaling (the obtained one after scaling the attack image is called the output image $\mathcal{O}$). One example of image-scaling attack is shown in Fig.~\ref{scaling_attack}, where an image of dog becomes the cat after performing the scaling step. The optimization problem of image-scaling attack can be formulated as:
\begin{align}
	& \ \widehat{\mathcal{A}} = \;\underset{\mathcal{A}}{\arg \min} \;d_{1}(\Delta_{1}) + c \cdot d_{2}(\Delta_{2}), \label{ori_optimization}\\
	  s.t.&
		\begin{cases}
		\Delta_{1} = \mathcal{S} - \mathcal{A}, \ \ \ \ \ \ \ \ \ \ \ \ \ \ \ \ \ \ \ \ \ \ \ \ \ \ \ \ \ \ \ \ \ (\ref{ori_optimization}a)\\
		\Delta_{2} = \mathcal{T} - \mathcal{O}, \ \ \ \ \ \ \ \ \ \ \ \ \ \ \ \ \ \ \ \ \ \ \ \ \ \ \ \ \ \ \ \ \ (\ref{ori_optimization}b)\\
		Pred(\mathcal{T}) = Pred(\mathcal{O}), \ \ \ \ \ \ \ \ \ \ \ \ \ \ \ \ \ \ \ \ \ \ \ (\ref{ori_optimization}c)\\
		\end{cases}\nonumber
	\end{align}
where $d_{1}(\cdot)$ and $d_{2}(\cdot)$ indicate the distance metrics, $c$ is the regulating parameter, $\mathcal{O} = {\rm ScaleFunc}(\mathcal{A}) = L \cdot \mathcal{A} \cdot R$, ${\rm ScaleFunc}(\cdot)$ denotes the scaling function, $L$ and $R$ represent the scaling matrices depending on the applied scaling algorithm~\cite{xiao2019}, $Pred(\cdot)$ is the prediction of the model.

From the optimization problem in Eq.~(\ref{ori_optimization}), image-scaling attack is expected to achieve the following two objectives:
\begin{itemize}
	\item \textbf{Objective I:} The obtained output images $\mathcal{O}$ after scaling the attack images $\mathcal{A}$ needs to be indistinguishable from the target images $\mathcal{T}$.
	\item \textbf{Objective II:} The crafted attack images $\mathcal{A}$ should be visually similar to the source images $\mathcal{S}$.
\end{itemize}

\section{Attack Mechanism}\label{sec:attack}
In this section, we first demonstrate the threat model in the attack. Then, we propose the attack mechanisms targeting on two scenarios, respectively.
 \subsection{Threat Model}
 ISP pipelines can transform RAW data into RGB images, which usually includes several processing subtasks to enhance the quality of the generated images. ISP pipeline applies some cascaded algorithm modules to address the subtasks, respectively. Define $h: \mathbb{R}^{d \times 1} \rightarrow \mathbb{R}^{d \times 3}$ as the ISP function that maps an RAW data $x \in \mathbb{R}^{d \times 1}$ to an RGB image $y \in \mathbb{R}^{d \times 3}$. Thus, the overall processing procedure of ISP pipeline can be represented as $y=h(x)=f_{N}\Big(f_{N-1}\big(\cdots f_{1}(x)\big)\Big)$, where $f_{i}(\cdot), 1\leq i \leq N$ indicates the $i$th algorithm module.

 In this work, an adversary focuses on tampering with the RAW data processed by ISP pipeline, which traditionally has not been considered susceptible to adversarial attacks, to launch the image-scaling attack. Concretely, the adversary manipulates the RAW data $\mathcal{S_{R}}$ to obtain adversarial RAW $\mathcal{A_{R}}$, which can reconstruct the attack image $\mathcal{A}$ through ISP pipeline. The obtained attack image $\mathcal{A}$ possesses a high similarity with the source image $\mathcal{S}$ generated from the clean RAW $\mathcal{S_{R}}$, making them visually indistinguishable. And the attack image $\mathcal{A}$ can be scaled to the target image $\mathcal{T}$. Considering two objectives of image-scaling attack and Eq.~(\ref{ori_optimization}), the optimization problem of image-scaling
 attack against ISP pipeline can be redefined as:
 \begin{align}
	& \ \widehat{\mathcal{\mathcal{A}}}_{\mathcal{R}} = \;\underset{\mathcal{\mathcal{A_{R}}}}{\arg \min} \;d_{1}(\Delta^{'}_{1}) + c \cdot d_{2}(\Delta^{'}_{2}), \label{ISP_optimization}\\
	  s.t.&
		\begin{cases}
		\Delta^{'}_{1} = \mathcal{S} - \mathcal{A} = h(\mathcal{S_{R}}) - h(\mathcal{A_{R}}),    \ \ \ \ \ \ \ \ \ \ \ \ \ \ (\ref{ISP_optimization}a)\\
		\Delta^{'}_{2} = \mathcal{T} - \mathcal{O}, \ \ \ \ \ \ \ \ \ \ \ \ \ \ \ \ \ \ \ \ \ \ \ \ \ \ \ \ \ \ \ \ \ \ \ \ \ (\ref{ISP_optimization}b)\\
		Pred(\mathcal{T}) = Pred(\mathcal{O}). \ \ \ \ \ \ \ \ \ \ \ \ \ \ \ \ \ \ \ \ \ \ \ \ \ \ \ (\ref{ISP_optimization}c)\\
		\end{cases}\nonumber
	\end{align}

Normally, the gradient-based approach is employed to solve the optimization problem in Eq.~(\ref{ISP_optimization}), i.e., to create the adversarial RAW data and launch the image-scaling attack. ISP pipeline acts as the `bridge' between RAW data and RGB images, and the gradient information of ISP pipeline can promote attack generations. To make the adversarial pattern applicable for each of the vision application scenarios, we investigate the following two adversarial attacks, the difference between which lies in whether the gradient information of the ISP pipeline can be obtained:
 \begin{itemize}
	\item \textbf{Gradient-available ISP attack:} It targets at the differentiable white-box ISP, in which the adversary can directly calculate the gradient information.
	\item \textbf{Gradient-unavailable ISP attack:} It targets at: 1) the non-differentiable white-box ISP, and 2) the black-box ISP. Note that the gradient information in either of the two cases is unavailable for the adversary.
\end{itemize}

 \subsection{Gradient-Available ISP Attack}
 Some known processing modules can constitute a simple ISP pipeline, so that the adversary can calculate their derivatives to get gradient information.
 Thus, we first consider a simple scenario, i.e., the gradient-available ISP pipeline. Since the gradient information is available, a given differentiable ISP pipeline can be directly utilized to launch an image-scaling attack. Based on Eq.~(\ref{ISP_optimization}), the objective function of gradient-available ISP attack can be defined as:
 \begin{align}
	\mathcal{L}_{obj}(h;\mathcal{A_{R}};\mathcal{S_{R}};\mathcal{T}) = \frac{1}{m \cdot n} \cdot ||\Delta^{'}_{1}||^{2}_{2} + c \cdot \frac{1}{m^{'} \cdot n^{'}} \cdot ||\Delta^{'}_{2}||^{2}_{2},
	\label{obj1}
 \end{align}
 where $\mathcal{L}_{2}$-norm metric is to calculate the similarity between two images, the factors $\frac{1}{m \cdot n}$ and $\frac{1}{m^{'} \cdot n^{'}}$ are adopted to reduce impacts of different dimensions, $m$ $(m^{'})$ and $n$ $(n^{'})$ represent the dimensions of the RAW data (the target image).

 In objective function Eq.~(\ref{obj1}), we select $\mathcal{L}_{2}$-norm as the distance metric for the following reasons: firstly, $\mathcal{L}_{2}$-norm is differentiable and can simplify the attack implementations; secondly, $\mathcal{L}_{2}$-norm is apt to cause the uniform distortions and can be more stable across distinct models. Overall, Eq.~(\ref{obj1}) is differentiable, which can be directly optimized by the gradient descent method.

\begin{algorithm}[H]
	\caption{Gradient-available ISP attack}
	\label{alg1}
	\begin{algorithmic}[1]
		\Require ISP function $h$; The RAW data $\mathcal{S_{R}}$; The target image $\mathcal{T}$; The regulating parameter $c$; Number of attack iterations $n$;  Learning rate $\alpha$.
		\Ensure	Adversarial RAW data $\mathcal{A_{R}}$.
		\State $\mathcal{S} = h(\mathcal{S_{R}})$; $\mathcal{A_{R}} = \mathcal{S_{R}}$;   $ \hfill \triangleright {\rm \ Initialization}$
		\For{$k \leftarrow 1 \cdots n$}    $ \hfill \triangleright {\rm \ Attack \ generation \ process}$
		\State $Grad_{\mathcal{A_{R}}} = \nabla_{\mathcal{A_{R}}}\mathcal{L}_{obj}(h;\mathcal{A_{R}};\mathcal{S_{R}};\mathcal{T})$;
		\State $\mathcal{A_{R}} = \mathcal{A_{R}} - Adam(\alpha, Grad_{\mathcal{A_{R}}})$; $ \hfill \triangleright {\rm \ Adam\ optimizer}$
		\EndFor
		\State $\mathcal{A_{R}} = clip(\mathcal{A_{R}})$;  $ \hfill \triangleright {\rm \ Clip\ the \ values \ to \ a\  valid\ range}$
		\State \Return{$A_{R}$};
	\end{algorithmic}
\end{algorithm}

\textbf{Adversarial RAW generation process:} We optimize Eq.~(\ref{obj1}) to perform the image-scaling attack targeting ISP pipeline. Through minimizing the objective function, the obtained adversarial RAW $\mathcal{A_{R}}$ can be converted into the attack image $\mathcal{A}$. We adopt the Adam optimizer to ensure the stability of the optimization process. Moreover, we exploit the clip operation to guarantee the values of adversarial RAW $\mathcal{A_{R}}$ fall into a valid range. The overall procedure of gradient-available ISP attack is presented in Algorithm~\ref{alg1}.

\begin{remark}
	Differentiable ISP pipeline often includes simple image processing modules, so high-quality RGB images cannot be reconstructed. To improve the performance of ISP pipeline, various complex processing modules are embedded, and even some are not open to users because of trade secrets. Therefore, obtaining gradient information to directly launch the attack is no longer possible. In the following part, we will present our solutions for gradient-unavailable ISP pipeline.
\end{remark}

 \subsection{Gradient-Unavailable ISP Attack}\label{Non-differentiable ISP attack}
  Notably, the high-performance ISP pipelines always consist of intricate processing modules, which are usually non-differentiable or black-box, then it is impossible to directly obtain the gradient information. Therefore, we further consider the gradient-unavailable ISP pipeline to make the proposed adversarial attacks more applicable. The gradient-unavailable ISP pipeline can be approximated by a proxy differentiable function, which maps the RAW data to RGB images via a CNN model. Distinct from the conventional ISP design, the CNN model utilizes the data-driven method and can combine multiple processing subtasks together.

 \textbf{Framework:} Utilizing the obtained image pairs for supervised learning, we propose an approximation model to effectively learn the transformation from RAW data to RGB images. We define a proxy differentiable function as $\tilde{h}: \mathbb{R}^{d\times 1} \rightarrow \mathbb{R}^{d \times 3}$ that approximates the gradient-unavailable ISP $h$. Given a set of RAW data $\mathcal{X}=\{x_{1},\cdots,x_{n}\}$ and the corresponding RGB images $\mathcal{Y}=\{y_{1},\cdots,y_{n}\}$ generated by $h$, the ultimate purpose is to train a proxy function $\tilde{h}: \mathcal{X} \rightarrow \mathcal{Y}$, such that, for a pair $x_{i} \in \mathcal{X}$ and $y_{i} \in \mathcal{Y}$, the reconstructed RGB image $\tilde{h}(x_{i})$ should match the target image $y_{i}$. Different from existing work, the proposed model directly consumes the RAW data without the Bayer pattern extracting procedure, making it uncomplicated to generate the adversarial RAW during optimization.

 \textbf{Loss functions:} We train our proxy model $\tilde{h}$ with the RAW data $\mathcal{X}$ and the target RGB images $\mathcal{Y}$ applying the content loss, indicated as $\mathcal{L}_{con}$, which measures $\mathcal{L}_{2}$-norm distance between the target RGB images and the reconstructed RGB images. $\mathcal{L}_{con}$ is presented as:
 \begin{align}
	\mathcal{L}_{con} = \mathbb{E}_{\mathcal{X}}||\mathcal{Y}-\tilde{h}(\mathcal{X})||^{2}_{2}. \nonumber
 \end{align}

 In addition to utilize the content loss to eliminate significant color deviations, we also apply the structural similarity (SSIM)~\cite{ssim} loss to enhance the dynamic range of the reconstructed images. The structural similarity of samples $a$ and $b$ is defined as:
 \begin{align}
	SSIM(a,b) = \frac{(2\mu_{a}\mu_{b}+c_{1})(2\sigma_{ab}+c_{2})}{(\mu^{2}_{a}+\mu^{2}_{b}+c_{1})(\sigma^{2}_{a}+\sigma^{2}_{b}+c_{2})},
	\nonumber
 \end{align}
 where $\mu_{a}$ and $\mu_{b}$ are the mean values, $\sigma_{a}$ and $\sigma_{b}$ indicate the variances, $\sigma_{ab}$ denotes the covariance, $c_{1}$ and $c_{2}$ represent the regulating constants.

 Thus, SSIM of the target images $\mathcal{Y}$ and the reconstructed RGB images $\tilde{h}(\mathcal{X})$ by the proxy model is denoted as:
 \begin{align}
	\mathcal{L}_{SSIM} = SSIM\left(\mathcal{Y},\tilde{h}(\mathcal{X})\right),
	\nonumber
 \end{align}
 where $\mathcal{L}_{SSIM}$ is within $[0,1]$ and a large value of SSIM indicates the high-quality reconstructed images.

 Perceptual distance is the image quality metric that extracts characteristics from pretrained perceptual network, which can reflect the semantic representations~\cite{per_loss}. To calibrate the semantic information of reconstructed RGB images, we utilize perceptual loss as follows:
 \begin{align}
	\mathcal{L}_{per} = \mathbb{E}_{\mathcal{X}}||w \odot \left(\phi\left(\mathcal{Y}\right)-\phi\left(\tilde{h}(\mathcal{X})\right)\right)||^{2}_{2},
	\nonumber
 \end{align}
 where $\phi(\cdot)$ is the results of the perceptual network, $w$ indicates the vector to scale the channel-wise activations.

 The total training loss of the proxy model is formulated as follows:
 \begin{align}
	\mathcal{L}_{total} = \mathcal{L}_{con} + \lambda_{1} \cdot (1 - \mathcal{L}_{SSIM}) + \lambda_{2} \cdot \mathcal{L}_{per},
	\label{total_loss}
 \end{align}
 where $\lambda_{1}$ and $\lambda_{2}$ are the hyper-parameters.

\begin{figure}[t]
	\centering
	\includegraphics[height=4.3cm,width=8.10cm]{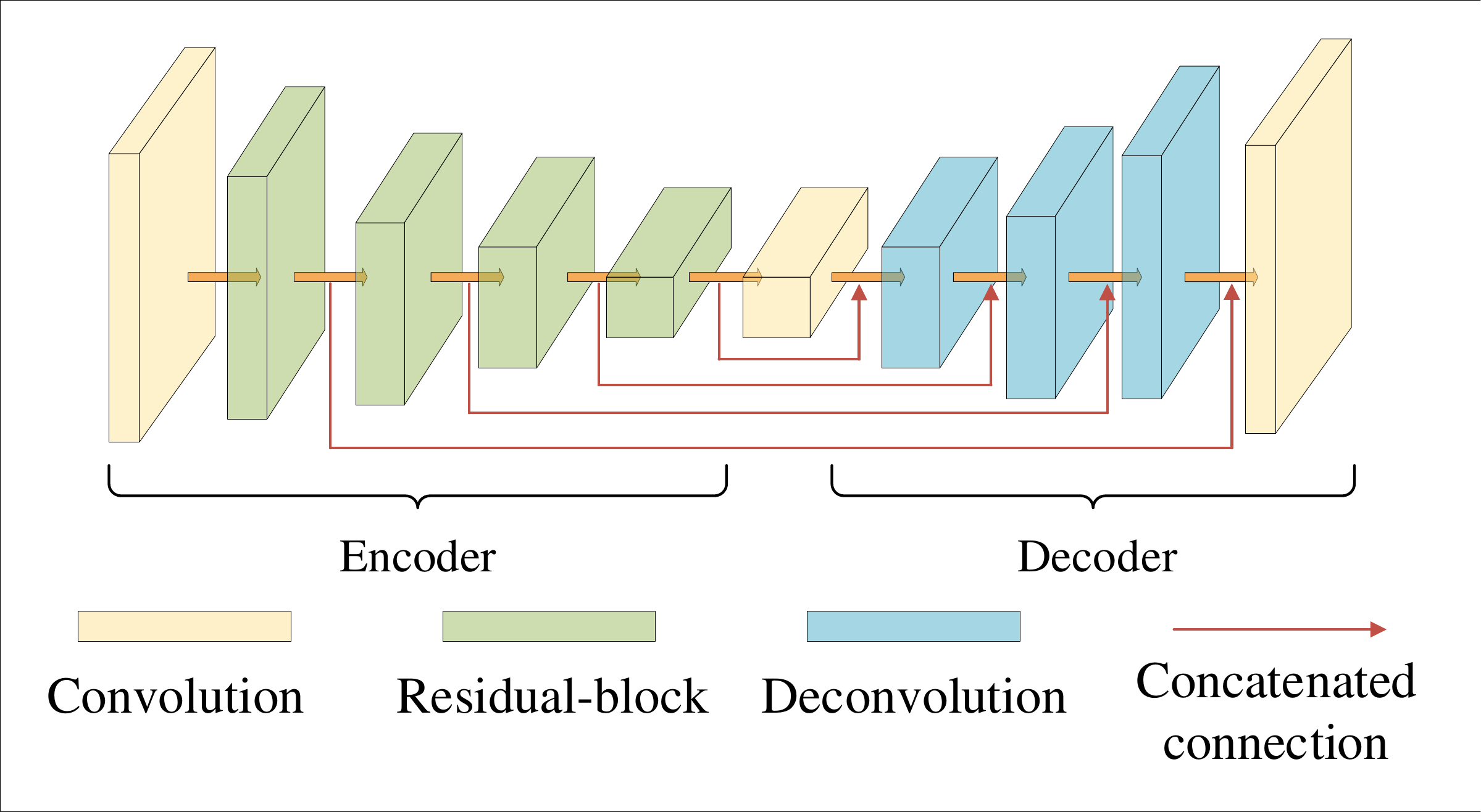}
	\caption{The network architecture of the proxy model, which consists of the encoder and decoder. The proxy model learns the RAW-to-RGB transformations.}
	\label{network}
\end{figure}

\begin{figure}[htbp!]
	\centering	
	\subfigure[]{
		\begin{minipage}[t]{0.5\linewidth}
			\centering
			\includegraphics[width=1.5in]{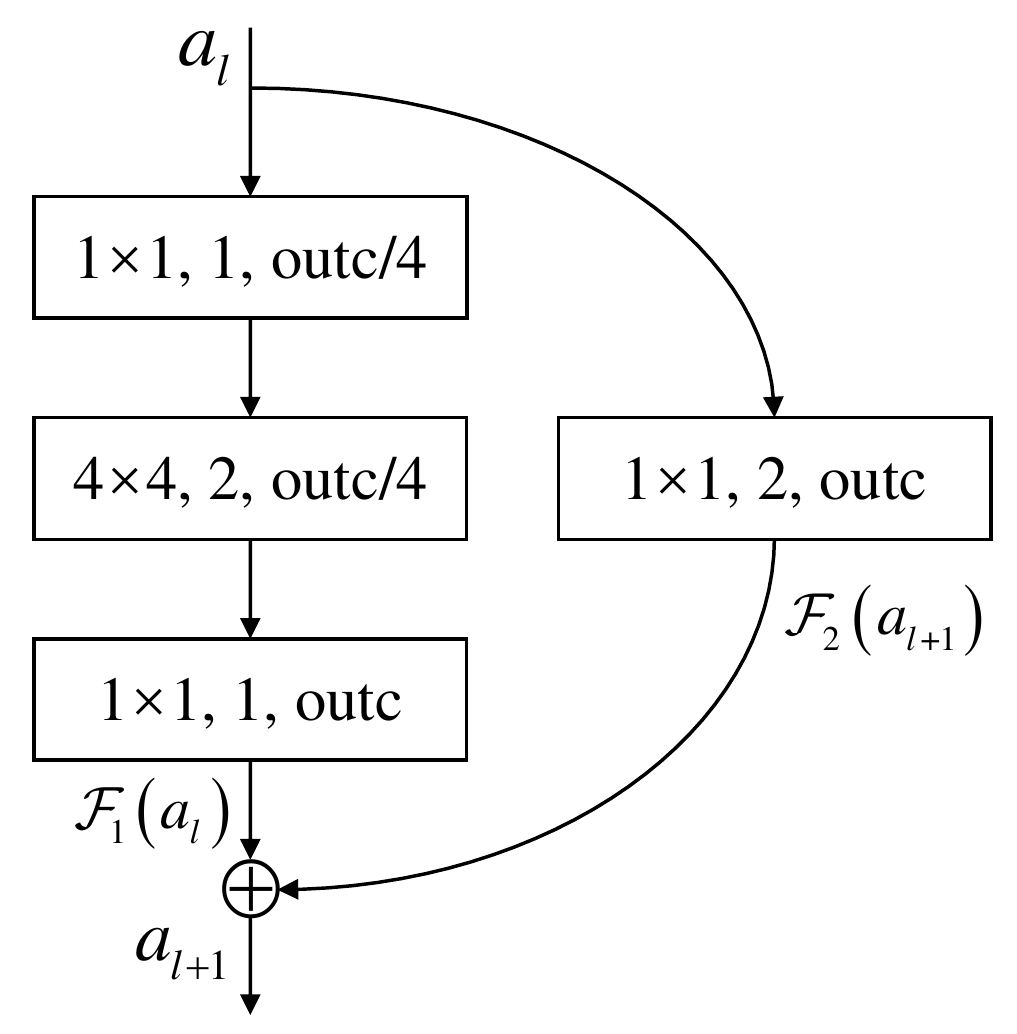}
		\end{minipage}%
	}%
	\subfigure[]{
		\begin{minipage}[t]{0.49\linewidth}
			\centering
			\includegraphics[width=1.5in]{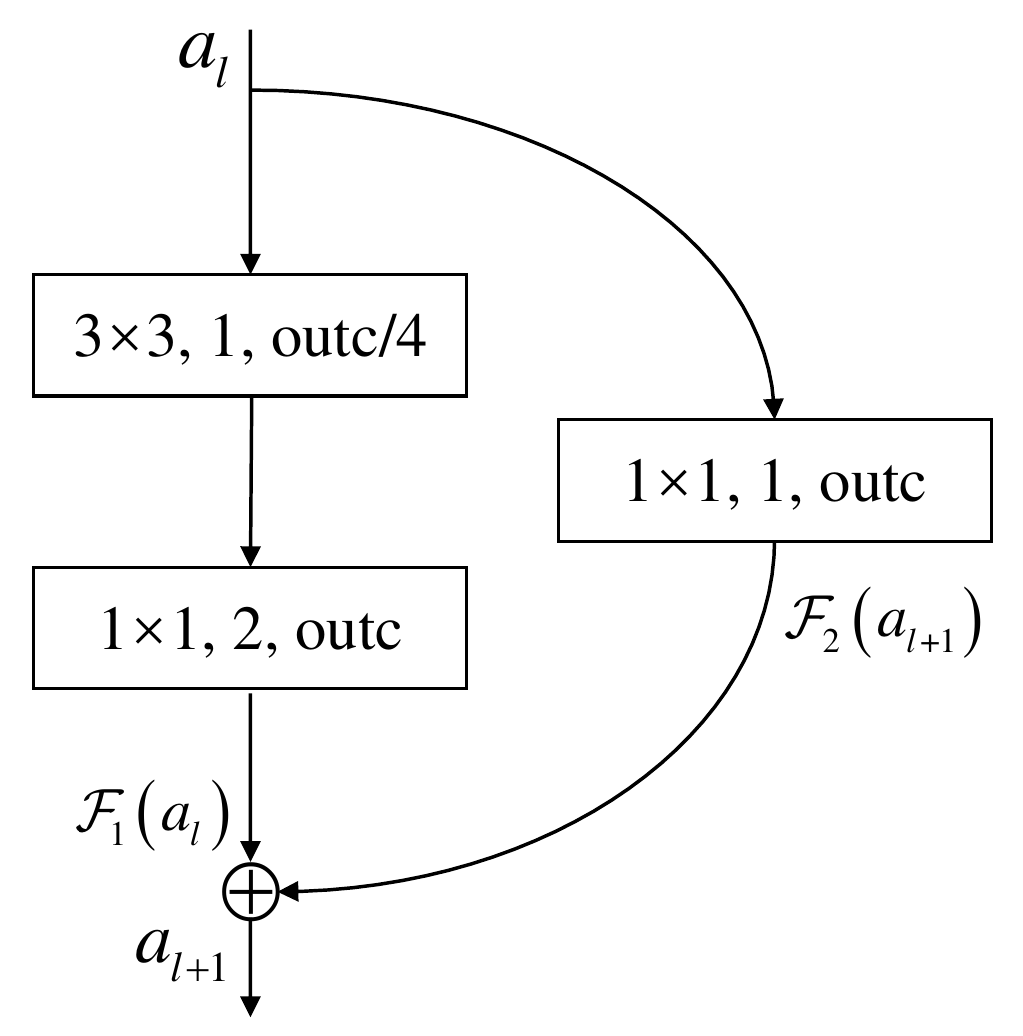}
		\end{minipage}%
	}%
	\centering
	\caption{Two structures of residual blocks used in the proxy model. The rectangles represent the convolutional operations, and the parameters in a rectangle denote the kernel size and stride.  The output channel is outc. $\mathcal{F}_{1}(a_{l})$ indicates output from the stack of convolution layers, and $\mathcal{F}_{2}(a_{l})$ is the output from the shortcut connection. The symbol $\oplus$ represents element-wised summation.}
	\label{residual_block}
\end{figure}

 \textbf{Network architecture:} In the proxy model, we use the encoding-decoding structure to effectively extract and reconstruct multi-level features. The proxy model is illustrated in Fig.~\ref{network}, where the encoder and decoder include convolution and deconvolution operations, respectively. The inputs are the RAW data, the outputs are the reconstructed RGB images that imitate the target RGB images. Inspired by \textit{DenseNet}~\cite{densenet}, we adopt the concatenated connections, i.e., the features from encoding layers are concatenated with their mirrored features generated by decoding layers, which can recover the low-level feature losses. To avoid the limitation of the input sizes, we only employ the convolutional and deconvolutional layers in the proxy model (except for the activation layers). Moreover, to improve the efficiency of low-level feature extractions, \textit{residual-block}~\cite{residual} consisting of multiple convolutional layers with a shortcut connection is used in the proxy model. Generally, a residual-block can be formulated as:
 \begin{align}
	a_{l+1} = \mathcal{F}_{1}(a_{l}) + \mathcal{F}_{2}(a_{l}),
	\nonumber
 \end{align}
where $\mathcal{F}_{1}(\cdot)$ and $\mathcal{F}_{2}(\cdot)$ indicate the residual functions, $a_{l}$ and $a_{l+1}$ are the input and the output of the residual-block, respectively. Fig.~\ref{residual_block} shows the two types of residual-blocks used in the proxy model.

\textbf{Training of the proxy model:} The training dataset is obtained by querying the target ISP pipeline. With the owned RAW data and the corresponding RGB images from the target ISP pipeline, the proxy model is trained by the supervised learning method. Since the convolutional and deconvolutional layers are differentiable, Adam optimizer is utilized to optimize the parameters of the proxy model. Once the model is trained well, the approximation of the gradient information can be applied to generate the adversarial RAW data.

\textbf{Adversarial RAW generation process:} We aim to craft the adversarial RAW data $\mathcal{A_{R}}$ that can generate the threatening attack image $\mathcal{A}$ by the target gradient-unavailable ISP pipeline. The generation process of the adversarial RAW data $\mathcal{A_{R}}$ is similar to Algorithm~\ref{alg1}. With the pretrained proxy model $\tilde{h}$, the gradient-based image-scaling attack can employ the approximated gradient to tamper with the clean RAW $\mathcal{S_{R}}$. Thus, the approximated gradient information can be obtained as:
\begin{align}
    Grad^{'}_{\mathcal{A_{R}}} = \nabla_{\mathcal{A_{R}}}\mathcal{L}_{obj}(\tilde{h};\mathcal{A_{R}};\mathcal{S_{R}};\mathcal{T}).\nonumber
\end{align}
$Grad^{'}_{\mathcal{A_{R}}}$ is as the approximation information of the adversarial RAW $\mathcal{A_{R}}$ in the attack generation. Then the updated $\mathcal{A_{R}}$ can be transferred well to the target ISP pipeline to generate the attack image $\mathcal{A}$ that can be scaled to the target image $\mathcal{T}$.

\begin{figure}[t]
	\centering
	\includegraphics[height=4.2cm,width=8.4cm]{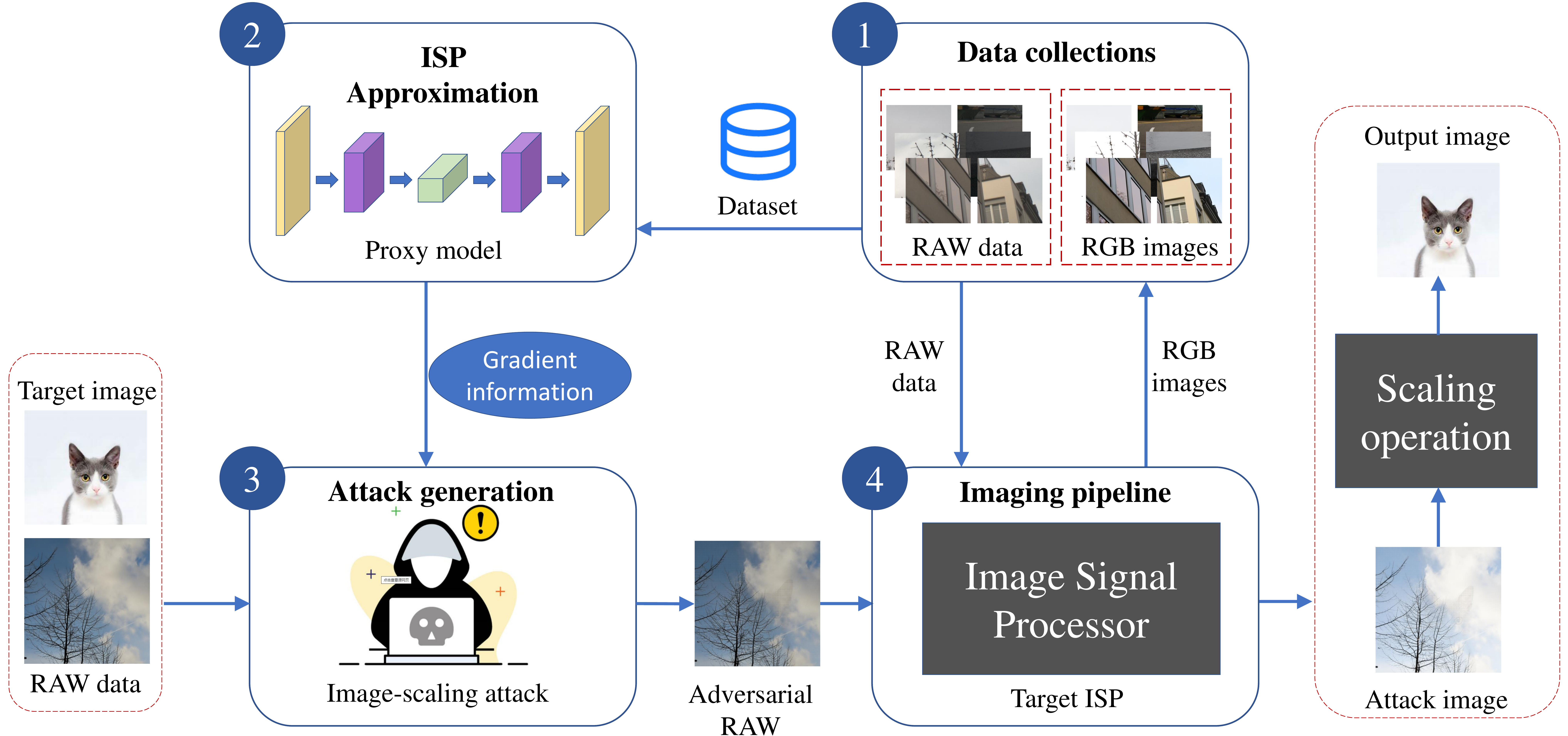}
	\caption{The overall procedure of image-scaling attack against gradient-unavailable ISP pipeline. (1) Collections of training data for the proxy model by querying the target ISP whose gradient information is unavailable. (2) Training the proxy model on the obtained RAW-RGB data pairs. (3) Adversarial RAW generation based on the proxy model. (4) Utilizing the crafted adversarial RAW to attack the target ISP pipeline and generate attack images.}
	\label{attack_process}
\end{figure}

The overall attack procedure is presented in Fig.~\ref{attack_process}, which includes: (1) Dataset collections: inputting the owned RAW data to the target ISP pipeline to obtain the corresponding RGB images; (2) Proxy model training: utilizing the obtained RAW-RGB data pairs to train the proxy modes as the approximation of the target ISP pipeline; (3) Attack generation: generating the adversarial RAW data by the gradient approximations from the proxy model; (4) Attack transferring: transferring the crafted adversarial RAW to the target ISP pipeline to obtain the attack images.

\section{Experimental Evaluation}\label{sec:experiment}
In this section, we evaluate the performance of the attack mechanisms. Firstly, we demonstrate the experimental setup. Then, we verify the image-scaling attack against both gradient-available and gradient-unavailable ISP pipelines. Finally, we extend the proposed attack to some current vision applications.
\subsection{Experimental Setup}
\textbf{Dataset:} We implement the experiments with the following datasets:
\begin{itemize}
	\item \textbf{Animals-10}~\cite{animal10}: It includes about 28K animal images belonging to ten categories. We use this dataset to train VGG-16 for evaluating the attack rates of the output images.
	\item \textbf{ImageNET}~\cite{imagenet}: It is a large visualization dataset for visual object recognition researches. We randomly select several images belonging to the categories of \textit{Animal10} from \textit{ImageNET} as the target images. Each image is adjusted to a shorter side length of 480 pixels while the aspect ratio of the image remains unchanged.
	\item \textbf{Zurich RAW to RGB dataset}~\cite{raw_data}: It is a large-scale dataset RAW patches for RAW to RGB mapping problems that consists of 20K photos. RAW patches (448 $\times$ 448) are utilized to obtain their corresponding RGB images and train the proxy model.
\end{itemize}

\textbf{ISP pipelines:} We evaluate the attack targeting on two gradient-available ISP pipelines: one performs bilinear demosaicing and the other implements bilateral filtering behind bilinear demosaicing (we call it bilateral filtering for short). Then, the attack for the gradient-unavailable OpenISP~\cite{openisp} is implemented.

\textbf{Scaling methods:} The used scaling algorithm can determine the distribution of the perturbations introduced into the RAW data. We select three common scaling methods (Nearest, Bilinear, Bicubic) in \textit{OpenCV}~\cite{opencv} for the evaluations. For the gradient-available ISP attack, we adopt Bilinear as the default scaling method. For the gradient-unavailable ISP attack, we utilize Nearest and Bicubic.

\textbf{Evaluation metrics:} A successful image-scaling attack should satisfy Objectives I and II. To assess Objective I, we check whether VGG-16 gives the identical results for the target image and the output image, called attack success rates (ASR). The quality of RGB images relies on RAW data and RAW data is not suitable for human eyes. To maintain the consistency of measure metrics, for Objective II, we observe the $\mathcal{L}_{2}$-norm loss between the original RAW data (source image) and the adversarial RAW data (attack image). The attack is considered successful if $\mathcal{L}_{2}$-norm loss of the source image and the attack image is below $0.0250$.

\subsection{Gradient-Available ISP Attack}
We first evaluate the attack for the gradient-available ISP pipelines. We utilize Canon EOS 6D to capture 86 RAW data. The corresponding source images can be generated by two chosen ISP pipelines. Through experiments and observations, we find that the attack success rates increase significantly when the regulating parameter $c$ in Eq.~(\ref{obj1}) is in the range [0.1, 10]. Finally, through further refining, we choose c to be 0.1, 0.3, 1.0, 2.5 and 10.

\begin{figure}[htbp!]
	\centering	
	\subfigure[Bilinear demosaicing]{
		\begin{minipage}[t]{0.5\linewidth}
			\centering
			\includegraphics[width=1.5in]{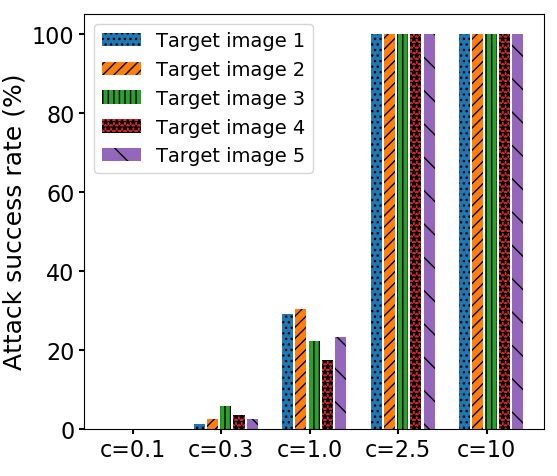}
		\end{minipage}%
	}%
	\subfigure[Bilateral filtering]{
		\begin{minipage}[t]{0.49\linewidth}
			\centering
			\includegraphics[width=1.5in]{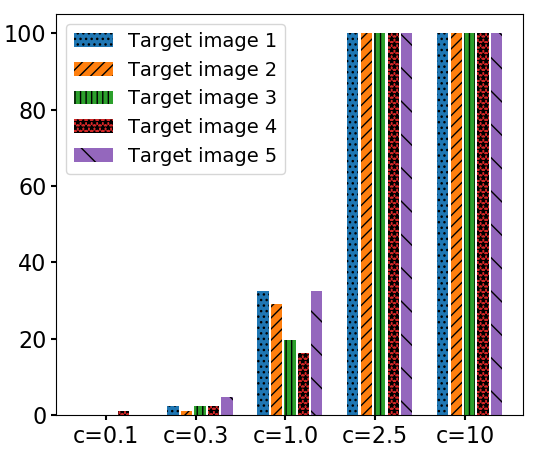}
		\end{minipage}%
	}%
	\centering
	\caption{Attack success rates targeting on two gradient-available ISP pipelines with different $c$.}
	\label{diff_success_rate}
\end{figure}

\textbf{Evaluation of Objective I:} We first give the experimental results regarding Objective I. Fig.~\ref{diff_success_rate} shows the attack performance corresponding to Objective I for two ISP pipelines, respectively. As $c$ increases, the attack success rates are also improved, i.e., the output images are increasingly indistinguishable from the target images. Especially when $c$ is set as 2.5 and 10, for all given target images, the proposed attacks can achieve 100$\%$ attack success rates. When $c$ is large, the more values in original RAW will be modified, so the obtained attack image will retain more contents of the target image after scaling. In other words, high attack success rates may expose the attack traces of attack images. Thus, we should further consider the similarity metrics between the source images and the attack images.

\begin{figure*}[t]
    \centering
	\includegraphics[height=12.04cm,width=18.0cm]{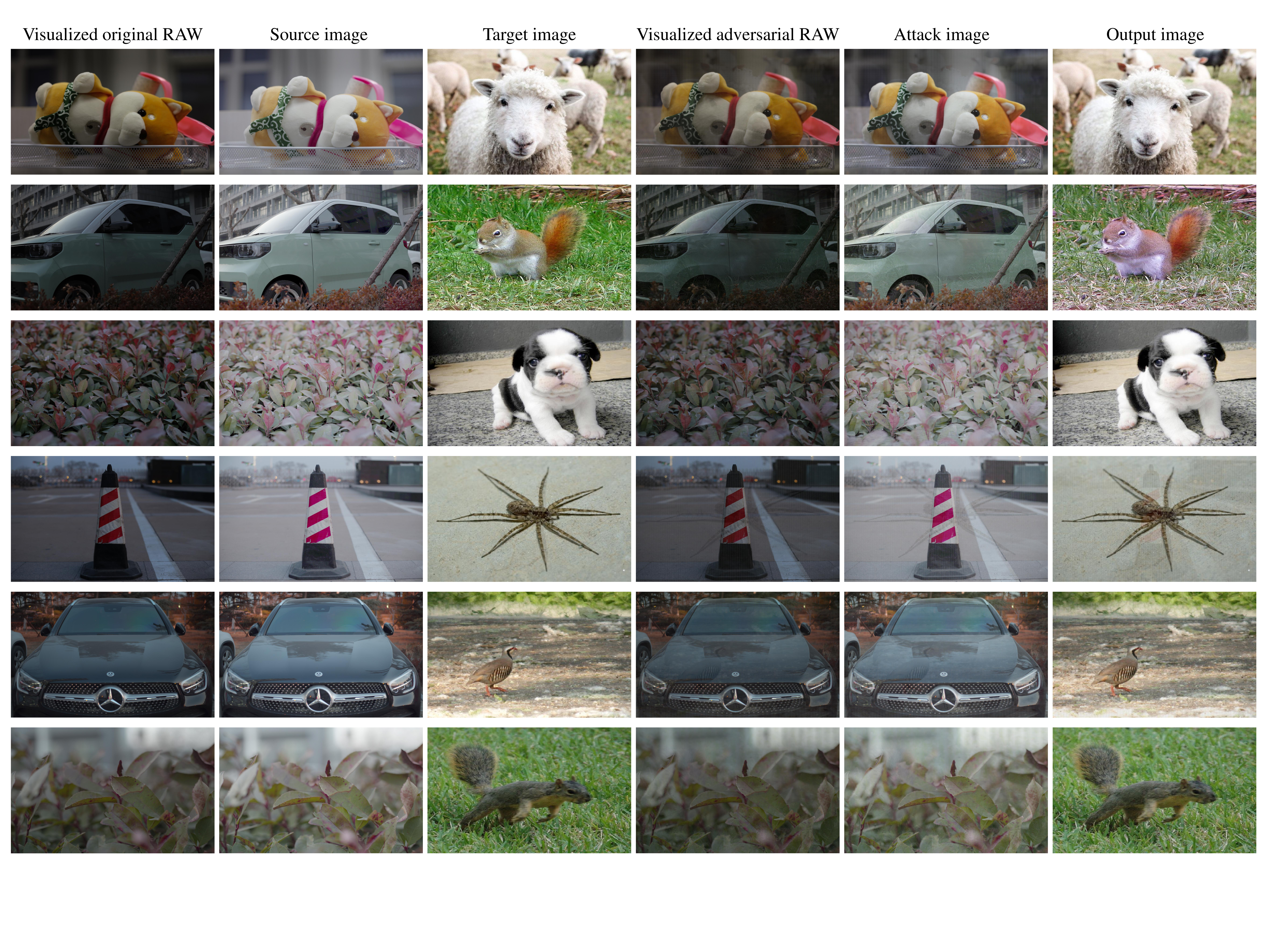}
    \caption{Examples of the image-scaling attack against the gradient-available ISP pipeline. The size of RAW data is 2736$\times$1824$\times$1, and each target image is resized to a shorter side length of 480 pixels while the
	aspect ratio of the image remains unchanged compared with its original sizes. For the sake of composition, we present the images (RAW data) as the same size.}
    \label{diff_attack_example}
\end{figure*}

\textbf{Evaluation of Objective II:} Then, we verify the results related to Objective II. Tables~\ref{bi_loss} and~\ref{fi_loss} show the loss changes for attacks when adjusting $c$. As can be seen, the losses between the original RAW and adversarial RAW are closed to the ones between the source images and the attack images, which demonstrates that the introduced attack patterns of adversarial RAW can be preserved by ISP pipelines. From Eq.~(\ref{obj1}), the key point is that the smaller losses between the original RAW and adversarial RAW mean the more deceiving attack images reconstructed by ISP pipelines. When increasing $c$, the losses become larger, i.e., the attack traces of attack images are more obvious. Through verifications and observations, when $c=2.5$, the obtained attack images can be deceiving as well as achieving high attack rates. Thus, $c=2.5$ can be exploited as the recommended setting for launching attacks.
\begin{table}[htbp!]
	\small
	\caption{Changes in losses of the image-scaling attack against bilinear demosaicing when adjusting $c$ in Eq.~(\ref{obj1}).}
	\label{bi_loss}
	\centering
	\renewcommand{\arraystretch}{1.25}
	\begin{tabular}{|c|ccccc|}
	\hline
	\multirow{2}{*}{\textbf{\begin{tabular}[c]{@{}c@{}}Target\\ image\end{tabular}}} & \multicolumn{5}{c|}{\textbf{Bilinear demosaicing}}                                                                             \\ \cline{2-6}
																					 & \multicolumn{1}{c|}{c=0.1\tablefootnote{For each target image when selecting a fixed $c$, the above item represents the loss between the original RAW and the adversarial RAW while the below item indicates the loss between the source image and the attack image.}}  & \multicolumn{1}{c|}{c=0.3}  & \multicolumn{1}{c|}{c=1.0}  & \multicolumn{1}{c|}{c=2.5}  & c=10   \\ \hline
	\multirow{2}{*}{1}                                                  & \multicolumn{1}{c|}{0.0051} & \multicolumn{1}{c|}{0.0148} & \multicolumn{1}{c|}{0.0176} & \multicolumn{1}{c|}{0.0236} & 0.0307 \\ \cline{2-6}
																					 & \multicolumn{1}{c|}{0.0048} & \multicolumn{1}{c|}{0.0134} & \multicolumn{1}{c|}{0.0171} & \multicolumn{1}{c|}{0.0233} & 0.0281 \\ \hline
	\multirow{2}{*}{2}                                                  & \multicolumn{1}{c|}{0.0059} & \multicolumn{1}{c|}{0.0155} & \multicolumn{1}{c|}{0.0196} & \multicolumn{1}{c|}{0.0223} & 0.0318 \\ \cline{2-6}
																					 & \multicolumn{1}{c|}{0.0061} & \multicolumn{1}{c|}{0.0131} & \multicolumn{1}{c|}{0.0197} & \multicolumn{1}{c|}{0.0217} & 0.0289 \\ \hline
	\multirow{2}{*}{3}                                                  & \multicolumn{1}{c|}{0.0072} & \multicolumn{1}{c|}{0.0169} & \multicolumn{1}{c|}{0.0205} & \multicolumn{1}{c|}{0.0255} & 0.0317 \\ \cline{2-6}
																					 & \multicolumn{1}{c|}{0.0072} & \multicolumn{1}{c|}{0.0155} & \multicolumn{1}{c|}{0.0201} & \multicolumn{1}{c|}{0.0241} & 0.0307 \\ \hline
	\multirow{2}{*}{4}                                                  & \multicolumn{1}{c|}{0.0056} & \multicolumn{1}{c|}{0.0112} & \multicolumn{1}{c|}{0.0147} & \multicolumn{1}{c|}{0.0211} & 0.0267 \\ \cline{2-6}
																					 & \multicolumn{1}{c|}{0.0044} & \multicolumn{1}{c|}{0.0089} & \multicolumn{1}{c|}{0.0141} & \multicolumn{1}{c|}{0.0196} & 0.0274 \\ \hline
	\multirow{2}{*}{5}                                                  & \multicolumn{1}{c|}{0.0047} & \multicolumn{1}{c|}{0.0126} & \multicolumn{1}{c|}{0.0201} & \multicolumn{1}{c|}{0.0220} & 0.0326 \\ \cline{2-6}
																					 & \multicolumn{1}{c|}{0.0050} & \multicolumn{1}{c|}{0.0119} & \multicolumn{1}{c|}{0.0185} & \multicolumn{1}{c|}{0.0215} & 0.0301 \\ \hline
	\end{tabular}
	\end{table}

\begin{table}[htbp!]
	\small
	\caption{Changes in losses of the image-scaling attack against bilateral filtering when adjusting $c$ in Eq.~(\ref{obj1}).}
	\label{fi_loss}
	\centering
	\renewcommand{\arraystretch}{1.25}
	\begin{tabular}{|c|ccccc|}
	\hline
	\multirow{2}{*}{\textbf{\begin{tabular}[c]{@{}c@{}}Target\\ image\end{tabular}}} & \multicolumn{5}{c|}{\textbf{Bilateral filtering}}                                                                              \\ \cline{2-6}
																						& \multicolumn{1}{c|}{c=0.1}  & \multicolumn{1}{c|}{c=0.3}  & \multicolumn{1}{c|}{c=1.0}  & \multicolumn{1}{c|}{c=2.5}  & c=10   \\ \hline
	\multirow{2}{*}{1}                                                                  & \multicolumn{1}{c|}{0.0038} & \multicolumn{1}{c|}{0.0093} & \multicolumn{1}{c|}{0.0155} & \multicolumn{1}{c|}{0.0197} & 0.0270 \\ \cline{2-6}
																						& \multicolumn{1}{c|}{0.0027} & \multicolumn{1}{c|}{0.0011} & \multicolumn{1}{c|}{0.0153} & \multicolumn{1}{c|}{0.0197} & 0.0252 \\ \hline
	\multirow{2}{*}{2}                                                                  & \multicolumn{1}{c|}{0.0048} & \multicolumn{1}{c|}{0.0131} & \multicolumn{1}{c|}{0.0205} & \multicolumn{1}{c|}{0.0275} & 0.0312 \\ \cline{2-6}
																						& \multicolumn{1}{c|}{0.0055} & \multicolumn{1}{c|}{0.0124} & \multicolumn{1}{c|}{0.0195} & \multicolumn{1}{c|}{0.0266} & 0.0290 \\ \hline
	\multirow{2}{*}{3}                                                                  & \multicolumn{1}{c|}{0.0042} & \multicolumn{1}{c|}{0.0171} & \multicolumn{1}{c|}{0.0212} & \multicolumn{1}{c|}{0.0208} & 0.0255 \\ \cline{2-6}
																						& \multicolumn{1}{c|}{0.0042} & \multicolumn{1}{c|}{0.0150} & \multicolumn{1}{c|}{0.0203} & \multicolumn{1}{c|}{0.0212} & 0.0237 \\ \hline
	\multirow{2}{*}{4}                                                                  & \multicolumn{1}{c|}{0.0068} & \multicolumn{1}{c|}{0.0127} & \multicolumn{1}{c|}{0.0186} & \multicolumn{1}{c|}{0.0241} & 0.0310 \\ \cline{2-6}
																						& \multicolumn{1}{c|}{0.0072} & \multicolumn{1}{c|}{0.0137} & \multicolumn{1}{c|}{0.0185} & \multicolumn{1}{c|}{0.0221} & 0.0289 \\ \hline
	\multirow{2}{*}{5}                                                                  & \multicolumn{1}{c|}{0.0047} & \multicolumn{1}{c|}{0.0162} & \multicolumn{1}{c|}{0.0205} & \multicolumn{1}{c|}{0.0222} & 0.0317 \\ \cline{2-6}
																						& \multicolumn{1}{c|}{0.0039} & \multicolumn{1}{c|}{0.0157} & \multicolumn{1}{c|}{0.0207} & \multicolumn{1}{c|}{0.0215} & 0.0295 \\ \hline
	\end{tabular}
	\end{table}

Examples of the gradient-available ISP attack is presented in Fig.~\ref{diff_attack_example}. Given the target images and gradient-available ISP pipelines, the crafted adversarial RAW data can be obtained to generate attack images. Once scaled to the specific sizes, those attack images can form the target contents.

\textbf{Summary:} From analysis and results above, we verify that a crafted adversarial RAW can be successfully against the gradient-available ISP pipeline. The available gradient information from ISP pipeline can promote constructions of the adversarial patterns properly. However, since most of the gradient-available ISP pipelines only support simple image processing, it is difficult for them to reconstruct high-quality RGB images.

\subsection{Gradient-Unavailable ISP Attack}
In this subsection, we demonstrate the results of the proposed methods against gradient-unavailable ISP pipeline.

\textbf{Proxy model training:} RAW patches of \textit{Zurich RAW to RGB dataset} are fed to the target ISP pipeline to obtain their corresponding RGB images. We can employ the RAW-RGB image pairs to train the proxy model. For efficient processing, we normalized the values of RAW data and RGB images to $[0,1]$.

\textbf{Attack generation:} After constructing the proxy model, we can craft the adversarial RAW against the target ISP pipeline. From the analysis above, we select the adjusting parameter $c=2.5$ to craft 100 adversarial RAW data. The attack generation procedure consists of two parts: Firstly, based on Algorithm~\ref{alg1}, we utilize the proxy model as the gradient oracle to generate the adversarial RAW. Then, the crafted adversarial RAW is transferred to OpenISP, i.e., the adversarial RAW can be transformed into the attack image by the target ISP pipeline.

\textbf{Evaluation of Objective I:} In Table~\ref{nondiff_asr}, the attack results targeting on OpenISP are demonstrated. For both two scaling methods, the generated attack images from OpenISP can achieve 100$\%$ attack success rates, i.e., their output images are mistaken by the classification model as the corresponding target images. The results illustrate that the proxy model simulates the processing procedure of the target ISP well. Meanwhile, the generated adversarial patterns of adversarial RAW through the proxy model can be transferred well to the target ISP to form the camouflaged contents after scaling. Though achieving high attack success rates, we still need to consider the loss metrics.
\begin{table}[htbp!]
	\small
	\caption{Attack success rates of gradient-unavailable ISP attack.}
	\label{nondiff_asr}
	\centering
	\renewcommand{\arraystretch}{1.25}
	\begin{tabular}{|c|ccccc|}
	\hline
	\multirow{2}{*}{\textbf{\begin{tabular}[c]{@{}c@{}}Scaling\\ method\end{tabular}}} & \multicolumn{5}{c|}{\textbf{Attack success rates ($\%$)}}                                                        \\ \cline{2-6}
																					   & \multicolumn{1}{c|}{1}   & \multicolumn{1}{c|}{2}   & \multicolumn{1}{c|}{3}   & \multicolumn{1}{c|}{4}   & 5   \\ \hline
	Nearest                                                                            & \multicolumn{1}{c|}{100} & \multicolumn{1}{c|}{100} & \multicolumn{1}{c|}{100} & \multicolumn{1}{c|}{100} & 100 \\ \hline
	Bicubic                                                                            & \multicolumn{1}{c|}{100} & \multicolumn{1}{c|}{100} & \multicolumn{1}{c|}{100} & \multicolumn{1}{c|}{100} & 100 \\ \hline
	\end{tabular}
	\end{table}

\textbf{Evaluation of Objective II:} Table~\ref{nondiff_loss} shows loss metrics of image-scaling attack against OpenISP. As can be seen, the losses between the adversarial RAW and original RAW are still similar to that between the attack images and source images, which again confirms that adversarial patterns of adversarial RAW are preserved by the target ISP. Generally, the losses between the source images and attack images are below $0.0250$, which means the generated attack images can be deceiving. Compared with Bicubic method, the adversarial RAW data (attack images) corresponding to Nearest generally have smaller losses given the same target image. Thus, the generated attack images corresponding to Nearest are more deceptive, which can hide the attack traces much better.
\begin{table}[h]
	\small
	\caption{Losses of the image-scaling attack against OpenISP with $c=2.5$.}
	\label{nondiff_loss}
	\centering
	\renewcommand{\arraystretch}{1.25}
	\begin{tabular}{|c|ccccc|}
	\hline
	\multirow{2}{*}{\textbf{\begin{tabular}[c]{@{}c@{}}Scaling\\ method\end{tabular}}} & \multicolumn{5}{c|}{\textbf{Target image}}                                                                                     \\ \cline{2-6}
																					   & \multicolumn{1}{c|}{1}      & \multicolumn{1}{c|}{2}      & \multicolumn{1}{c|}{3}      & \multicolumn{1}{c|}{4}      & 5      \\ \hline
	\multirow{2}{*}{Nearest}                                                           & \multicolumn{1}{c|}{0.0139} & \multicolumn{1}{c|}{0.0117} & \multicolumn{1}{c|}{0.0194} & \multicolumn{1}{c|}{0.0073} & 0.0143 \\ \cline{2-6}
																					   & \multicolumn{1}{c|}{0.0125} & \multicolumn{1}{c|}{0.0091} & \multicolumn{1}{c|}{0.0151} & \multicolumn{1}{c|}{0.0070} & 0.0111 \\ \hline
	\multirow{2}{*}{Bicubic}                                                           & \multicolumn{1}{c|}{0.0271} & \multicolumn{1}{c|}{0.0220} & \multicolumn{1}{c|}{0.0251} & \multicolumn{1}{c|}{0.0182} & 0.0223 \\ \cline{2-6}
																					   & \multicolumn{1}{c|}{0.0272} & \multicolumn{1}{c|}{0.0209} & \multicolumn{1}{c|}{0.0247} & \multicolumn{1}{c|}{0.0151} & 0.0207 \\ \hline
	\end{tabular}
	\end{table}

\begin{figure*}[t]
	\centering
	\includegraphics[height=12.04cm,width=18.0cm]{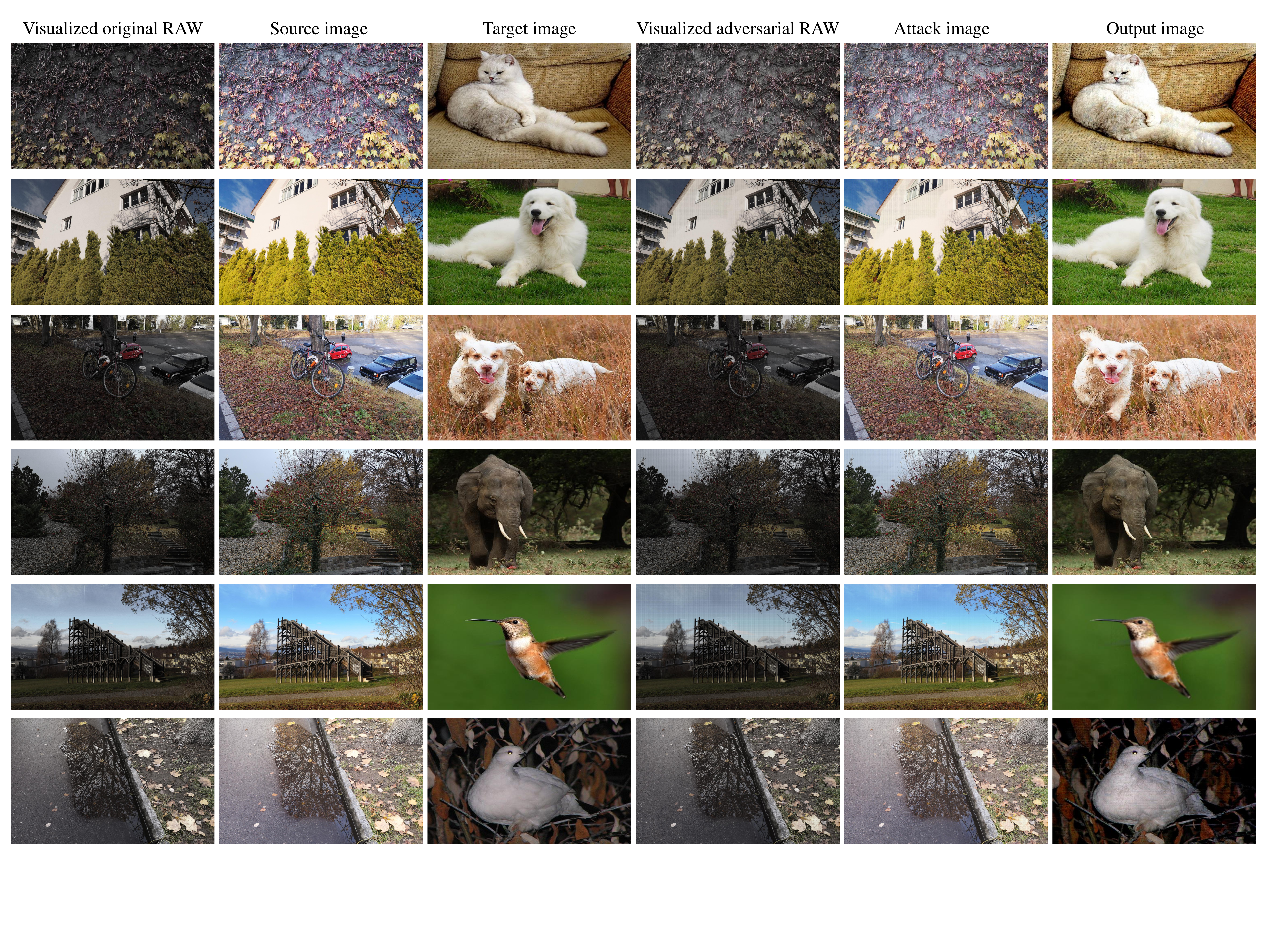}
	\caption{Examples of the image-scaling attack against the gradient-unavailable ISP pipeline. The sizes of RAW data include 2800$\times$1800$\times$1 and 4800$\times$3100$\times$1.}
	\label{no_diff_attack_example}
\end{figure*}

Examples of gradient-unavailable ISP attack are illustrated in Fig.~\ref{no_diff_attack_example}. The adversarial RAW data generated by the approximated gradient information of the proxy model can be transformed by the target ISP pipeline to the attack images successfully. The attack images still retain the adversarial contents from the target images.

\textbf{Summary:} Although the gradient information is unknown, the proxy model can utilize RAW-RGB pairs from the target ISP pipeline to approximate the gradient. We can not only utilize the proxy model to manipulate the original RAW but effectively transfer the generated adversarial RAW to the gradient-unavailable ISP pipeline.

\subsection{Attack Against Vision Applications}
\begin{figure}[htbp]
		\centering
		\includegraphics[height=6.79cm,width=8.5cm]{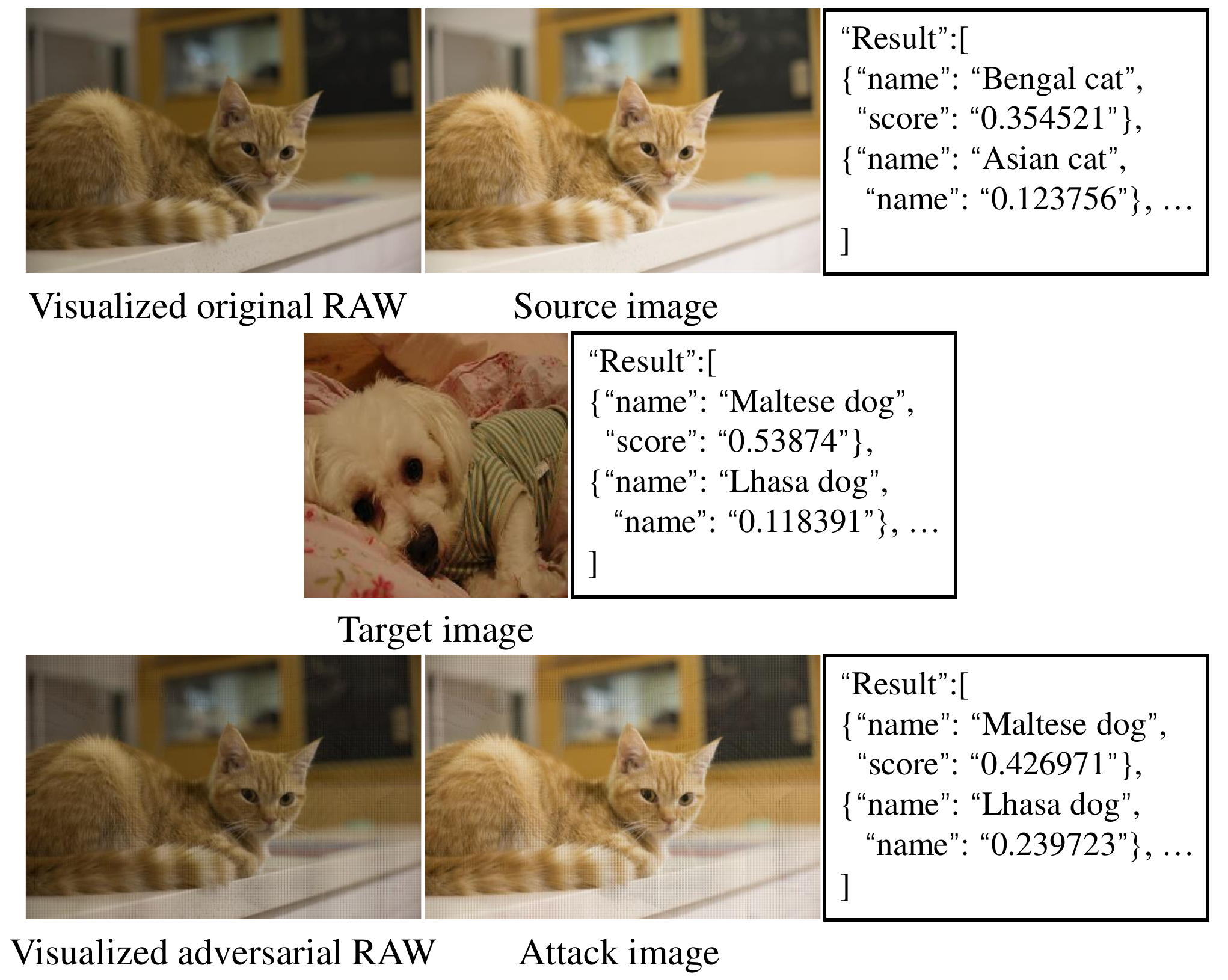}
		\caption{Example of adversarial data against Baidu animal classification, where Top-1 and Top-2 prediction results are presented. The size of RAW data is 1400$\times$900$\times$1, and the target image gets resized to 256$\times$256$\times$3. Notably, the predictions of Baidu animal classification are more specific. Thus, we consider the predicted results belonging to the target broad categories, such as `dog' or `cat', are all the
		expected predictions.}
		\label{baidu_animal}
	\end{figure}

\begin{figure}[h]
		\centering
		\includegraphics[height=4.4cm,width=8.41cm]{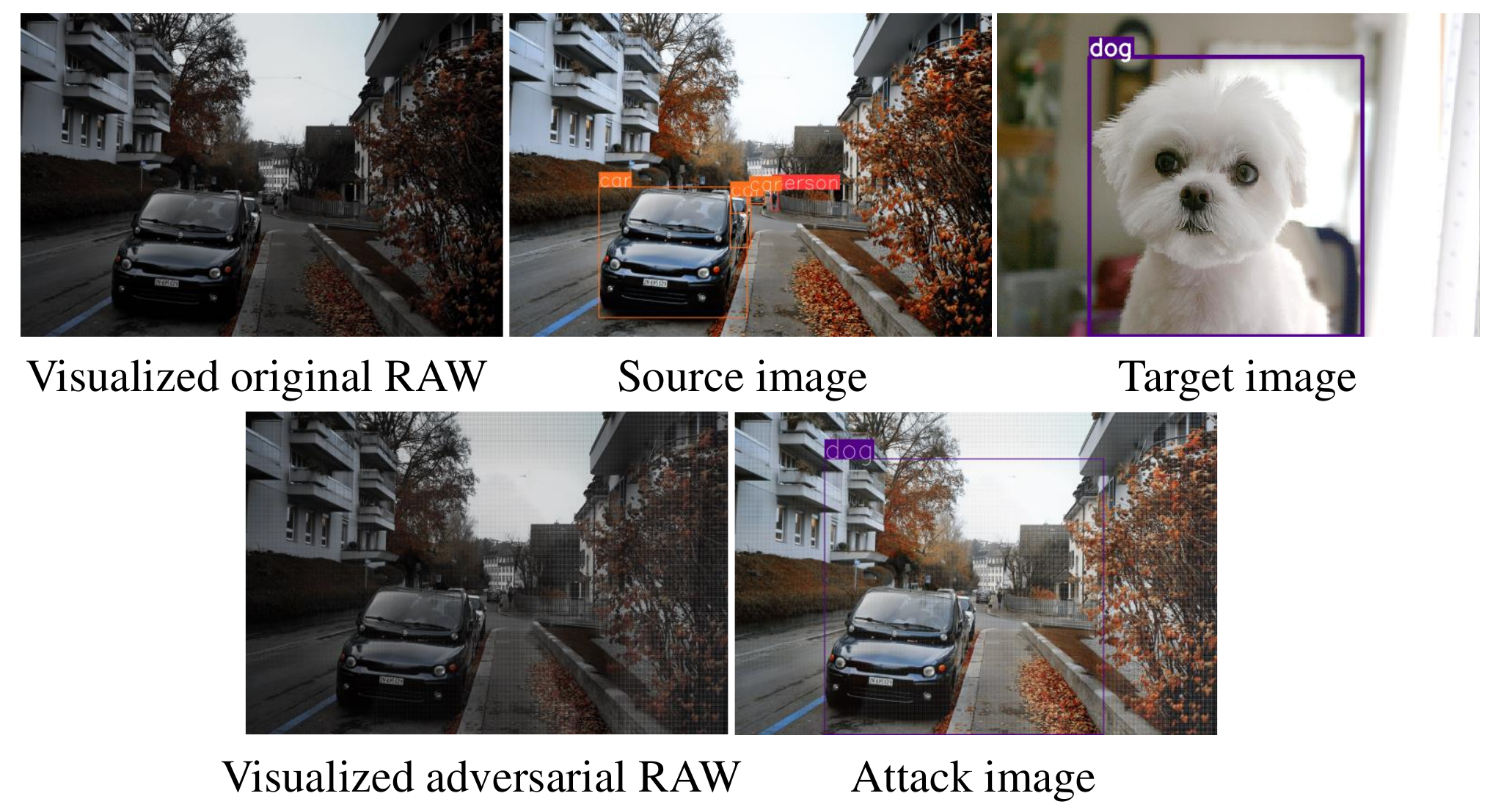}
		\caption{Example of adversarial data against Yolo-$v$5. The size of RAW data is 2800$\times$1800$\times$1, and the target image gets resized to 640$\times$411$\times$3 regarding the scaling rule of Yolo-$v$5.}
		\label{attack_yolo}
	\end{figure}

We further extend the proposed attack to be against some current vision applications as shown in Table~\ref{application_effects}, in which the applications are assumed to follow the OpenISP. We select three target images and ensure that their classifications are different from the generated source images. Considering the classification applications, each target image is adjusted to fit the input sizes of models. For the Yolo-$v$5 detection application, we resize the target images on the basis of its scaling rule. Moreover, the regulating parameter $c$ is set as 2.5. In each test, 50 adversarial RAW data are generated to obtain attack images. As shown in Table~\ref{application_effects}, the proposed attacks can achieve 100$\%$ attack success rates for classification applications. For Yolo-$v$5, the crafted attack images are tagged with the classifications of the target images. The results in Table~\ref{application_effects} further demonstrate the enormous threats to the practical vision applications by the adversarial attack. Fig.~\ref{baidu_animal} gives the attack results for Baidu animal classification, i.e., the attack images obtained from the adversarial RAW is mistaken for the expected class. An attack example against the pretrained Yolo-$v$5 is shown in Fig.~\ref{attack_yolo}, where the generated attack image is recognized as the classification of the target image.
\begin{table}[htbp]
	\small
	\caption{Attack effects on vision applications, where targets 1, 2 and 3 represent the three target images.}
	\label{application_effects}
	\centering
	\renewcommand{\arraystretch}{1.25}
	\begin{tabular}{|c|c|ccc|}
	\hline
	\multirow{2}{*}{\textbf{\begin{tabular}[c]{@{}c@{}}Vision\\ application\end{tabular}}}     & \multirow{2}{*}{\textbf{\begin{tabular}[c]{@{}c@{}}Interpolation\\ Method\end{tabular}}} & \multicolumn{3}{c|}{\textbf{Attack success rates (\%)}}                                                                        \\ \cline{3-5}
																							   &                                                                                          & \multicolumn{1}{c|}{Target 1}                       & \multicolumn{1}{c|}{Target 2}                      & Target 3                      \\ \hline
	\multirow{2}{*}{\begin{tabular}[c]{@{}c@{}}VGG-19\\ (224$\times$224)\end{tabular}}         & NEAREST                                                                                  & \multicolumn{1}{c|}{100}                      & \multicolumn{1}{c|}{100}                     & 100                     \\ \cline{2-5}
																							   & LINEAR                                                                                   & \multicolumn{1}{c|}{100}                      & \multicolumn{1}{c|}{100}                     & 100                     \\ \hline
	\multirow{2}{*}{\begin{tabular}[c]{@{}c@{}}ResNet\\ (224$\times$224)\end{tabular}}         & NEAREST                                                                                  & \multicolumn{1}{c|}{100}                      & \multicolumn{1}{c|}{100}                     & 100                     \\ \cline{2-5}
																							   & CUBIC                                                                                    & \multicolumn{1}{c|}{100}                      & \multicolumn{1}{c|}{100}                     & 100                     \\ \hline
	\multirow{2}{*}{\begin{tabular}[c]{@{}c@{}}Inception-$v$3\\ (299$\times$299)\end{tabular}} & NEAREST                                                                                  & \multicolumn{1}{c|}{100}                      & \multicolumn{1}{c|}{100}                     & 100                     \\ \cline{2-5}
																							   & CUBIC                                                                                    & \multicolumn{1}{c|}{100}                      & \multicolumn{1}{c|}{100}                     & 100                     \\ \hline
	\begin{tabular}[c]{@{}c@{}}Baidu animal\\ classification\\ (256$\times$256)\end{tabular}   & CUBIC                                                                                    & \multicolumn{1}{c|}{100}                      & \multicolumn{1}{c|}{100}                     & 100                     \\ \hline
	Yolo-$v$5(640)                                                                             & LINEAR                                                                                   & \multicolumn{3}{c|}{\begin{tabular}[c]{@{}c@{}}All attack images are \\ tagged with the target results.\end{tabular}} \\ \hline
	\end{tabular}
	\end{table}

\section{Discussions On Defense Directions}\label{sec:defense}
Essentially, the proposed attack modifies some points in RAW data, and the pixels converted from these points are preserved to form target contents when scaling. Thus, the defense method should restore those adversarial points in RAW data, i.e., to reconstruct `clean' points belonging to the `semantics' of the source image, which is similar to well-known problem in image processing technologies, i.e., image reconstruction. The filtering methods are often utilized to eliminate the perturbations in images. Thus, we analyse the defense effects of the two following typical filtering methods for the proposed attack:
\begin{itemize}
	\item \textbf{Average filtering:} Given an image, there are a set of pixels $\mathcal{P}$. For each pixel $p \in \mathcal{P}$, average filtering determines a window $W_{a}$ around $p$ and computes the average pixel value for this window to replace $p$.
	\item \textbf{Median filtering:} Median filtering computes the median pixel value for a window $W_{m}$ around $p$ to replace $p$.
\end{itemize}

Given the scaling method and the target ISP pipelines, we generate the adversarial RAW data to obtain the attack images. Then, we utilize the two filtering methods to process the adversarial RAW data. Both of the two filtering methods can reduce attack success rates to 0, i.e., those adversarial points in RAW data are successfully destroyed and `semantics' of the images converted by ISP pipelines no longer belong to the target images. However, through further experiments, we found that although images converted from the processed RAW data are not recognized by the model as the categories of the target images, some generated images are also not correctly classified, i.e., their original `semantics' are not recovered. Actually, the window of the filtering methods may cover both the `clean' points and the adversarial points of RAW data. Thus, the `clean' points may also be affected by the adversarial points or other `clean' points when implementing filtering process, which may lead to the losses of the original `semantics' of the generated images. Notably, the aim of defense is not only to restore the adversarial points in RAW data but also to recover the `semantics' of the generated images. Therefore, these two filtering methods are not necessarily reliable means for defense. Ideally, the defense method should precisely capture the adversarial points in RAW data and restore them without compromising the `clean' points, which greatly ensures that the original 'semantics' of the generated images are preserved. Thus, the more effective and reliable strategies should be further investigated to defend against the proposed attack.

\textbf{Summary:} The typical filtering methods can destroy the adversarial points of RAW data but may not render the original `semantics' of the generated RGB images be recovered. The more effective defense strategies should be developed to ensure the defense effects.

\section{Conclusions And Future Works}\label{sec:conclusion}
In this paper, we study the image-scaling attack against ISP pipeline, in which the generated images from ISP pipeline can cause remarkable changes of image `semantics' after scaling by tampering with RAW data. We first consider the gradient-available ISP pipeline, in which the gradient information can be directly used to generate the adversarial RAW to launch the attack. To make the adversarial attack more applicable, we further consider the gradient-unavailable ISP pipeline, in which a proxy model that well learns the RAW-to-RGB transformations is proposed as the gradient oracles. We conduct extensive experiments to validate the effectiveness of the proposed attacks. Finally, we analyse the defense effects of two common filtering methods and point out the the future direction of defense strategies.

We have investigated the potential attacks on ISP pipeline. In the future, we will further explore vulnerabilities of ISP pipeline and develop appropriate defense measures to strengthen the security.
\ifCLASSOPTIONcaptionsoff
  \newpage
\fi
\bibliographystyle{IEEEtran}
\bibliography{references}

\end{document}